\documentclass[11pt]{article}

\usepackage{acl}

\usepackage{times}
\usepackage{latexsym}
\usepackage{amsmath}

\usepackage[T1]{fontenc}

\usepackage[utf8]{inputenc}

\usepackage{microtype}

\usepackage{inconsolata}

\usepackage{graphicx}
\usepackage{booktabs}
\usepackage{colortbl}
\usepackage{afterpage}
\newcommand{\downchange}[2]{$#1_{\textcolor{green!50!black}{\scriptscriptstyle\downarrow#2}}$}
\newcommand{\upchange}[2]{$#1_{\textcolor{red}{\scriptscriptstyle\uparrow#2}}$}
\title{Attend, Transform, or Silence: Operator-Level Visual Skipping for Efficient
Multimodal LLM Inference}

\author{
 \textbf{Zhaoyang Luo}\textsuperscript{1},
 \textbf{Runmin Dong}\textsuperscript{2}\textsuperscript{*},
 \textbf{Miao Yang}\textsuperscript{4},
 \textbf{Fan Wei}\textsuperscript{4},
\\
 \textbf{Yushan Lai}\textsuperscript{1},
 \textbf{Bin Luo}\textsuperscript{1},
 \textbf{Haohuan Fu}\textsuperscript{1,3,4}\textsuperscript{*},
\\
 \textsuperscript{1}Tsinghua Shenzhen International Graduate School, Shenzhen, China \\
 \textsuperscript{2}Sun Yat-sen University, Zhuhai, China \\
 \textsuperscript{3}National Supercomputing Center in Shenzhen, Shenzhen, China \\
 \textsuperscript{4}Tsinghua University, Beijing, China \\
\\
 Correspondence: \href{mailto:luocy25@mails.tsinghua.edu.cn}{\nolinkurl{luocy25@mails.tsinghua.edu.cn}}; \href{mailto:dongrm3@mail.sysu.edu.cn}{\nolinkurl{dongrm3@mail.sysu.edu.cn}}
}

\begin{document}
\maketitle
\begingroup
\renewcommand{\thefootnote}{*}
\begin{NoHyper}\footnotetext{Corresponding authors.}\end{NoHyper}
\endgroup
\begin{abstract}
Multimodal large language models (MLLMs) increasingly process long visual-token sequences, increasing the overall inference computation. Existing acceleration methods usually remove visual tokens or skip visual-token updates in entire layers, but these coarse strategies may discard fine-grained evidence or suppress useful operators together with redundant ones. In this paper, we study visual-token computation from an answer-observable perspective and find that late visual-token updates can remain large while having little effect on answer-token representations. Motivated by this answer-silent redundancy, we decompose each Transformer layer into attention and FFN operators and show that useful visual computation is often operator-dominant and layer-dependent. We propose an operator-level visual-token skipping framework that preserves the full visual-token sequence while selectively bypassing redundant attention, FFN, or both. Experiments across three MLLM architectures and 10 VQA benchmarks show that our method achieves strong efficiency-accuracy trade-offs, reducing \textbf{33.7\%} TFLOPs on Qwen3-VL while retaining \textbf{99.5\%} of the vanilla model performance.\footnote{The source code
is available at: \url{https://github.com/zayan-l/Op-Skip}.}

\end{abstract}

\section{Introduction}
\label{sec:introduction}
\begin{figure}[!t]
    \centering
    \includegraphics[width=\columnwidth]{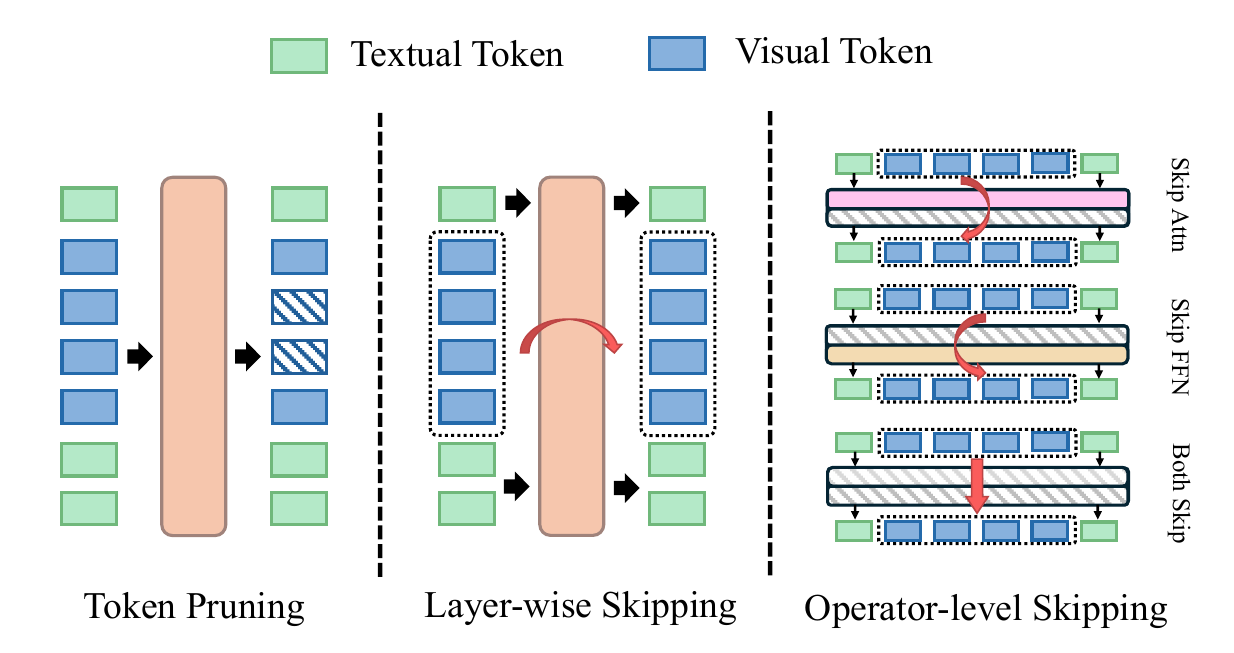}
    \caption{
    Overview of visual-computation reduction granularities for MLLM inference.
    Token pruning removes selected visual tokens, which disrupts spatial relationships and causes irreversible loss of visual evidence needed for reasoning.
    Layer-wise skipping preserves the token sequence but treats each Transformer layer as an indivisible unit, failing to fully exploit the distinct roles of attention and FFN operators.
    Our operator-level skipping selectively skips attention, FFN, or both while preserving the full visual-token sequence.
    }
    \label{fig:teaser}
    \vspace{-1em}
\end{figure}

Multimodal large language models (MLLMs)~\citep{llava15,llavaonevision,llava-next,qwen25vl,qwen3vl} have become a central paradigm for visual understanding, visual reasoning, and visual question answering. A typical MLLM encodes an image into visual tokens, projects them into the embedding space of a large language model (LLM), and concatenates them with text tokens for autoregressive generation~\citep{llava15,qwen25vl,qwen3vl}. To preserve fine-grained visual evidence, recent MLLMs increasingly adopt high-resolution inputs, image tiling, or dense visual encoders, expanding the visual sequence from hundreds to thousands of tokens~\citep{blip2,llava-next,qwen25vl,qwen3vl}. While this improves perceptual coverage, it also makes the prefill stage expensive, as the LLM backbone must process all visual and textual tokens simultaneously through self-attention and feed-forward transformations~\citep{fastv}. A growing body of work therefore accelerates MLLMs by reducing visual computation, either by pruning, selecting, or merging visual tokens~\citep{nuwa,pdrop,cdpruner,visionzip}, or by skipping visual-token updates in selected Transformer layers~\citep{shortv,vtw,vskip}. These methods show that not all visual computation is equally necessary, but they rely on coarse notions of redundancy. Token-level methods may permanently discard fine-grained evidence, layer-level skipping removes attention and FFN computation together, and V-Skip~\citep{vskip} assumes a fixed preference for preserving FFN while skipping visual attention. Such fixed granularities cannot decide whether attention, FFN, or both are redundant at each layer, and may suppress useful computation when only part of a layer is redundant.

In this paper, we revisit visual redundancy in MLLMs by examining how visual tokens evolve inside the LLM backbone. We begin with a counterintuitive observation, in late layers, visual-token hidden states can still change substantially, even when these changes have little influence on the final answer. This indicates that update magnitude alone is not a reliable measure of useful visual computation. To distinguish representational movement from computation that actually affects answer generation, we introduce a set of \emph{answer-observable} diagnostics, which measure how visual-token updates propagate to answer-token hidden states. Across multiple MLLM architectures and VQA benchmarks, we consistently find that late visual tokens often undergo large hidden-state updates while exhibiting small answer-observable effects. These updates are therefore not inactive, but \emph{answer-silent}, they consume computation while contributing little to the representations that determine the generated response.

This observation suggests that visual redundancy is not merely a property of individual tokens, nor is it always aligned with entire Transformer layers. Instead, redundancy may emerge at a finer operator level within each layer. This view is supported by recent analyses of LVLM backbones~\citep{lost_in_attention}, which reveal a functional decoupling between attention and FFN operators, attention primarily reconfigures information within a relatively preserved representation subspace, whereas FFNs expand the representation subspace and drive semantic transformation. If these two operators serve different representational roles, their importance for visual tokens should not be assumed to be coupled in every MLLM layer. Some layers may require attention to route visual evidence toward the question, some may rely more on FFN transformations to refine visual semantics, and others may contribute little through either operator.

Motivated by this observation, we decompose each Transformer layer into its attention and FFN operators and ask a finer question: \emph{which operator is actually necessary for visual-token computation in each layer?} By separately measuring the answer-observable contribution of attention and FFN, we find strong layer-wise heterogeneity, some layers are attention-dominant, some are FFN-dominant, and others show little answer-observable contribution. Based on this observation, we propose an operator-level visual-token skipping framework that preserves the full visual-token sequence while bypassing redundant visual computation. This design avoids the irreversible evidence loss of token pruning and releases visual skipping from the static assumption that either attention or FFN is always the important operator. Experiments across multiple MLLM architectures and 10 visual question answering benchmarks show that our framework substantially reduces visual computation during inference while maintaining answer quality, achieving up to \textbf{33.7\%} visual-computation reduction with only \textbf{0.5\%} average performance change. Beyond efficiency, our analysis suggests that useful visual computation in MLLMs is operator-dependent rather than block-aligned, and that efficient inference should be designed around answer-observable computation rather than raw update magnitude or fixed architectural boundaries.
\begin{figure*}[!t]
    \centering
    \includegraphics[width=\textwidth]{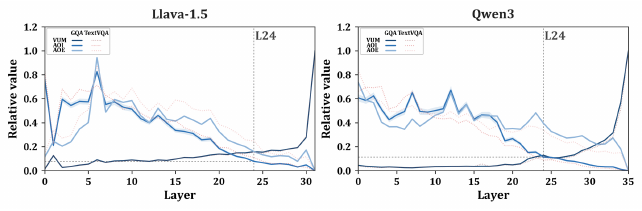}
    \caption{
    Layer-wise visual update magnitude, answer-observable influence, and answer-observable efficiency.
    All curves are normalized within each model and dataset for visualization.
    Across both GQA and TextVQA, late layers exhibit increasing visual-token update magnitude but decreasing answer-observable influence and efficiency.
    }
    \label{fig:answer_silent_curves}
\end{figure*}
Our contributions are summarized as follows:
{
\begin{itemize}
    \item We identify answer-silent visual-token updates in late MLLM layers and introduce answer-observable diagnostics to measure whether visual computation actually propagates to answer-token representations.
    \item We reveal that useful visual computation is operator-dominant and layer-dependent by decomposing each Transformer layer into attention and FFN operators, exposing visual redundancy beyond token- and layer-level views.
    \item We propose an operator-level visual-token skipping framework that preserves the full visual-token sequence while bypassing redundant visual computation, saving up to \textbf{33.7\%} visual computation while retaining \textbf{99.5\%} model performance across Qwen3VL and 10 VQA benchmarks.
\end{itemize}}

\section{Answer-Silent Visual Updates in MLLM}

\label{sec:answer_silent_updates}

We first examine when visual-token information is effectively used by the textual stream during the prefill stage. In mainstream decoder-only MLLMs~\citep{llava15,qwen25vl,qwen3vl}, visual tokens are placed before the textual prompt and processed under causal masking. As a result, the last prompt token is the first position that can integrate both the visual prefix and the textual question. Its hidden state, which we refer to as the \emph{final prompt-token hidden state}, is directly projected by the language-modeling head to produce the next-token logits. This motivates the following diagnostic question:

\noindent{\setlength{\fboxsep}{4pt}\fcolorbox{violet!55}{violet!8}{\parbox{\dimexpr\columnwidth-2\fboxsep-2\fboxrule\relax}{\textbf{Question 1.} \emph{Do visual-token updates remain answer-observable across all layers?}}}}

Let $H_v^l \in \mathbf{R}^{N_v \times d}$ denote the visual-token hidden states at layer $l$, where $N_v$ is the number of visual tokens and $d$ is the hidden dimension. We first quantify the magnitude of visual-token state changes between consecutive layers:
\begin{equation}
\resizebox{0.96\columnwidth}{!}{$\displaystyle
\Delta V_l = H_v^{l+1} - H_v^l,\quad
\mathrm{VUM}_l =
\frac{|\Delta V_l|_F}{\max_j |\Delta V_j|_F + \epsilon}.
$}
\label{eq:vum}
\end{equation}
We refer to this normalized quantity as visual update magnitude (VUM), where larger values indicate stronger visual-state changing. 

To measure whether such changing is observable from the answer side, we estimate its first-order influence on the final prompt-token hidden state. Let $h_{\mathrm{last}}$ denote the final prompt-token hidden state during prefill, and let
\begin{equation}
J_l =
\frac{\partial h_{\mathrm{last}}}{\partial H_v^{l+1}}
\end{equation}
be the Jacobian that maps perturbations of visual-token states at layer $l+1$ to changes in $h_{\mathrm{last}}$. We define
\begin{equation}
\delta h_l = J_l[\Delta V_l],
\qquad
\mathrm{AOI}_l =
|\delta h_l|_2 .
\label{eq:aoi}
\end{equation}
and we call $\mathrm{AOI}_l$ answer-observable influence (AOI), where $J_l[\Delta V_l]$ is the Jacobian-vector product estimating how the visual update at layer $l$ changes the final prompt-token hidden state.

Finally, we normalize AOI by the size of the visual update:
\begin{equation}
\resizebox{0.98\columnwidth}{!}{$\displaystyle
\mathrm{AOE}_l =
\frac{\mathrm{AOI}_l}{\|\Delta V_l\|_F + \epsilon}\;
\widetilde{\mathrm{AOE}}_l =
\frac{\mathrm{AOE}_l}{\max_j \mathrm{AOE}_j + \epsilon}.
$}
\label{eq:aoe}
\end{equation}
We refer to this quantity as answer-observable efficiency (AOE). In summary, VUM measures the magnitude of visual-token changes, AOI measures how much these changes reach the final prompt-token hidden state, and AOE measures the answer-observable influence per unit of visual-token update.

\begin{figure*}[!t]
    \centering
    \includegraphics[width=\textwidth]{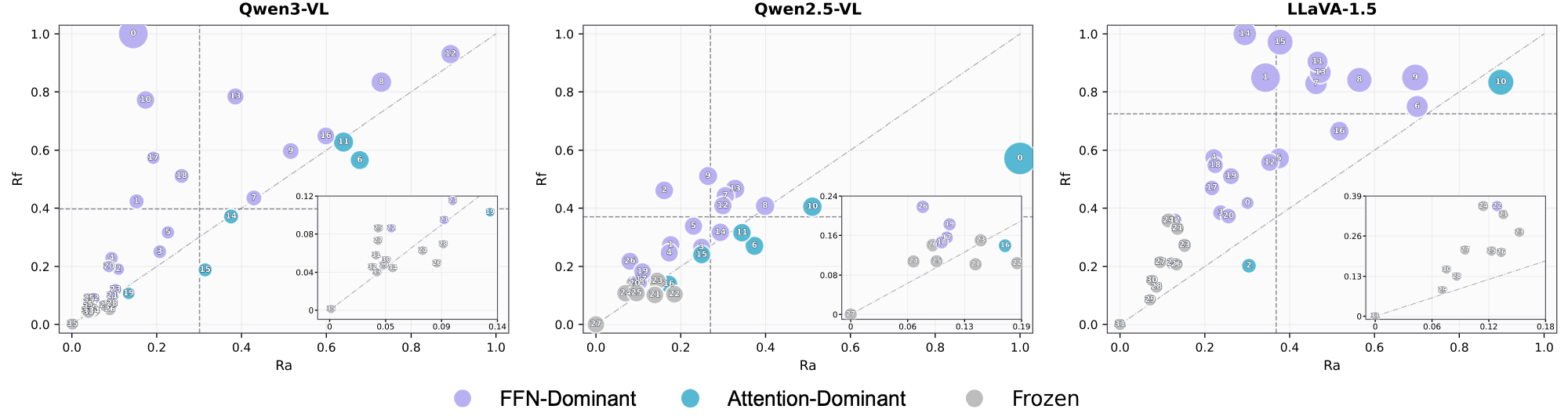}
    \caption{
    Layer-wise operator-risk analysis across Qwen3-VL, Qwen2.5-VL, and LLaVA-1.5.
    Each point denotes one Transformer layer, with the horizontal axis measuring the risk of skipping visual attention and the vertical axis measuring the risk of skipping the visual FFN.
    Layers above the diagonal are FFN-domain layers, layers below the diagonal are attention-domain layers, and low Raf late layers are assigned to the frozen domain, inset plots zoom in on low-risk regions where frozen layers concentrate.
    }
    \label{fig:layer_operator_analysis}
    \vspace{-0.8em}
\end{figure*}

Figure~\ref{fig:answer_silent_curves} shows the layer-wise behavior of VUM, AOI, and AOE on GQA and TextVQA for LLaVA-1.5 and Qwen3-VL. The curves reveal a consistent late-layer decoupling: visual-token states continue to change substantially, while their answer-observable influence decreases. In LLaVA-1.5, VUM rises sharply after layer 24, whereas both AOI and AOE continue to decline. Qwen3-VL exhibits the same pattern around its decoupling onset layer. Thus, raw update magnitude systematically overestimates the usefulness of late visual computation.

This observation suggests a direct intervention. If late visual-token updates are weakly coupled to answer generation, freezing visual-token states after the decoupling onset layer should have limited impact on task performance. We therefore freeze visual-token hidden states after layer 24 while keeping all text-token computation unchanged. Concretely, for layers $l > 24$, we reuse the visual states from layer 24 and allow textual tokens to continue attending to these fixed visual representations. This removes late visual-state changes while preserving the visual evidence accessible to textual tokens.

\begin{table}[t]
    \centering
    \caption{
    Effect of freezing late visual-token states.
    We freeze visual-token hidden states after the decoupling 24 layer while keeping text-token computation unchanged.
    }
    \label{tab:freeze_late_visual}
    \resizebox{\columnwidth}{!}{%
    \begin{tabular}{clcccc}
        \toprule
        Model & Method & GQA & POPE & TextVQA & SQA \\
        \midrule
        \raisebox{-0.5\normalbaselineskip}{LLaVA-1.5-7B} & Full & 61.9 & 85.9 & 58.2 & 69.5 \\
        & Frozen visual & 61.1$_{\textcolor{green}{\downarrow0.8}}$ & 86.4$_{\textcolor{red}{\uparrow0.5}}$ & 57.2$_{\textcolor{green}{\downarrow1.0}}$ & 69.6$_{\textcolor{red}{\uparrow0.1}}$ \\
        \midrule
        \raisebox{-0.5\normalbaselineskip}{Qwen3-VL-8B} & Full & 61.6 & 89.1 & 80.0 & 94.4 \\
        & Frozen visual & 61.3$_{\textcolor{green}{\downarrow0.3}}$ & 88.9$_{\textcolor{green}{\downarrow0.2}}$ & 80.1$_{\textcolor{red}{\uparrow0.1}}$ & 94.4 \\
        \bottomrule
    \end{tabular}}
\end{table}

As shown in Table~\ref{tab:freeze_late_visual}, freezing late visual-token states causes only marginal degradation on 4 benchmarks across the two MLLM architectures. This result indicates that repeatedly updating visual-token states is often not the computation that determines next-token prediction. The model can still access visual evidence through the textual stream, while the continued changes of late visual-token states become increasingly answer-silent.

\vspace{0.3em}
\noindent{\setlength{\fboxsep}{5pt}\fcolorbox{orange!65!brown}{yellow!14}{\parbox{\dimexpr\columnwidth-2\fboxsep-2\fboxrule\relax}{\textbf{Finding 1:} Late-layer visual tokens can change substantially while becoming increasingly answer-silent, freezing them removes considerable visual computation with limited performance loss.}}}
\vspace{0.1em}

This finding motivates a finer-grained analysis of where useful visual computation resides within each Transformer block. We therefore move beyond the block-level view and decompose each layer into its attention and FFN operators, analyzing which operator dominates answer-observable visual computation at different depths.

\section{Methodology}

\begin{figure*}[!t]
    \centering
    \includegraphics[width=\textwidth]{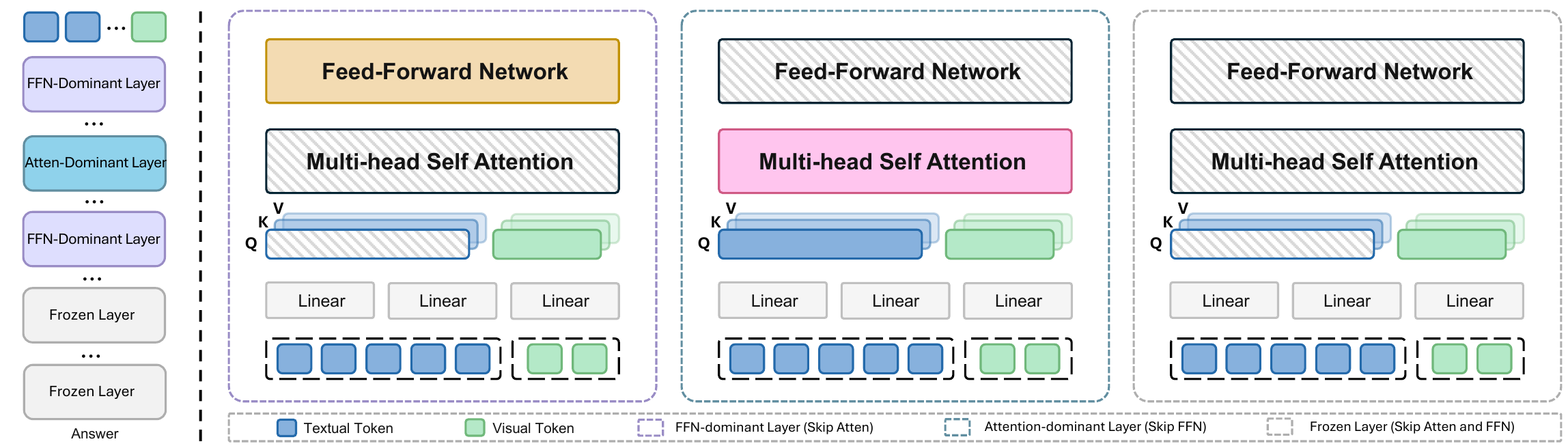}
    \caption{
    Overview of the proposed operator-level visual-token skipping policy.
    Each layer is assigned to an operator domain according to its visual-token contribution, FFN-domainant layers preserve visual-token FFN computation while skipping redundant visual-token attention, attention-domainant layers preserve visual-token attention while skipping redundant visual-token FFN computation, and frozen layers skip both visual-token attention and FFN updates.
    Textual tokens continue to follow the original Transformer computation, while the full visual-token sequence is retained throughout inference.
    }
     \vspace{-0.8em}
    \label{fig:operator_policy}
\end{figure*}

\subsection{Operator Risk Analysis}
\label{sec:operator-risk}

We first describe how to identify the layer-wise operator structure of the visual branch without retraining the model or using human-annotated answers. Given a decoder-only MLLM, let the hidden states at layer $l$ be
\begin{equation}
H^l = [H_{\mathcal V}^l; H_{\mathcal T}^l].
\end{equation}
where $H_{\mathcal V}^l \in \mathrm{R}^{N_v \times d}$ denotes the visual-token hidden states and $H_{\mathcal T}^l$ denotes the textual hidden states, including the system prompt, question tokens, and possible answer-prefix tokens. We decompose the visual branch of a Transformer block into two residual updates:
\begin{equation}
\begin{aligned}
\bar{H}_{\mathcal{V}}^{l}
&= H_{\mathcal{V}}^{l}
+ \big[A^{l}(H^{l})\big]_{\mathcal{V}}, \\
H_{\mathcal{V}}^{l+1}
&= \bar{H}_{\mathcal{V}}^{l}
+ \big[F^{l}(\bar{H}^{l})\big]_{\mathcal{V}} .
\end{aligned}
\label{eq:operator_decomposition}
\end{equation}
where $A^{l}$ is the self-attention update and $F^{l}$ is the
FFN update at layer $l$. The subscript $\mathcal{V}$ denotes the visual-token rows. This decomposition allows us to ask a more fine-grained question than whether an entire layer can be skipped: 

\vspace{0.3em}
\noindent{\setlength{\fboxsep}{4pt}\fcolorbox{violet!55}{violet!8}{\parbox{\dimexpr\columnwidth-2\fboxsep-2\fboxrule\relax}{\textbf{Question 2.} \emph{Which operator dominates the visual-branch computation at each layer?}}}}
\vspace{0.1em}

For an input sample $x$, the full model produces a reference next-token distribution from the final prompt-token hidden state after prefill:
\begin{equation}
p_0(\cdot|x)=\mathrm{softmax}(z_0(x)).
\end{equation}
This reference distribution is obtained from the model itself and does not require ground-truth labels. We estimate layer-wise operator risks on a small calibration set $\mathcal{D}$ sampled from multiple benchmarks, using these samples only to probe how the output distribution changes under counterfactual visual-operator interventions.
\begin{equation}
\begin{aligned}
&\frac{1}{|\mathcal{D}|}
\sum_{x\in\mathcal{D}}
D_{\mathrm{KL}}\!\left(
 p_0(\cdot|x)\Vert p_{s}^{(l)}(\cdot|x)
\right), \\
&\quad s\in\{\mathcal A,\mathcal F,\mathcal{AF}\}.
\end{aligned}
\end{equation}

Here, $\mathcal{R}{\mathcal A}(l)$, $\mathcal{R}{\mathcal F}(l)$, and $\mathcal{R}_{\mathcal{AF}}(l)$ measure the output risk of removing visual attention, removing visual FFN, and freezing the visual branch at layer $l$, respectively.

\begin{equation}
\log
\frac{\mathcal{R}{\mathcal A}(l)+\epsilon}
{\mathcal{R}{\mathcal F}(l)+\epsilon}.
\end{equation}
A positive $\Gamma_l$ indicates an attention-dominant layer, while a negative $\Gamma_l$ indicates an FFN-dominant layer. Layers with uniformly small risks, especially small $\mathcal{R}_{\mathcal{AF}}(l)$, are treated as frozen layers.

\vspace{0.3em}
\noindent{\setlength{\fboxsep}{5pt}\fcolorbox{orange!65!brown}{yellow!14}{\parbox{\dimexpr\columnwidth-2\fboxsep-2\fboxrule\relax}{\textbf{Finding 2:} Visual computation in MLLM Transformer layers exhibits layer-dependent operator dominance, suggesting that redundancy reduction should be guided by the dominant operator of each layer.}}}
\vspace{-0.6em}

\subsection{Operator-Aware Skipping}
\label{sec:operator-aware-skipping}

We convert the layer-wise risk profile into an inference-time skipping policy through two steps, budgeted layer selection and policy assignment. Given a compute budget $B$, we rank layers by $\mathcal{R}{\mathcal A}(l)+\mathcal{R}{\mathcal F}(l)$ and select the $B$ lowest-risk layers for simplification. These are layers where removing either visual attention or visual FFN individually induces only a small output shift.

For each selected layer, we assign the least risky visual-branch policy. If $\mathcal{R}{\mathcal A}(l) < \mathcal{R}{\mathcal F}(l)$, we apply \textsc{FFN-only}, skipping visual attention while preserving visual FFN. If $\mathcal{R}{\mathcal F}(l) < \mathcal{R}{\mathcal A}(l)$, we apply \textsc{Attention-only}, preserving visual attention while skipping visual FFN. If $\mathcal{R}{\mathcal{AF}}(l)$ is sufficiently small, we apply the more aggressive \textsc{Freeze-update} policy, skipping both visual operators and directly propagating $H{\mathcal V}^l$ to the next layer.

Unlike token pruning, our method preserves the full visual-token sequence and avoids irreversible evidence removal. Unlike layer-level skipping, it does not treat attention and FFN as an inseparable block, but preserves the dominant visual operator while bypassing redundant visual-branch computation.

\begin{table*}[t]
    \centering
    \caption{
    Main comparison across three MLLM architectures and 10 multimodal benchmarks. Avg. Ret. denotes the mean score ratio to the corresponding vanilla baseline
across benchmarks. Red and blue indicate the best and second-best results among methods within each architecture.
    }
    \label{tab:main_results}
    \resizebox{\textwidth}{!}{
    \begin{tabular}{lllccccccccccc}
        \toprule
        Model & Method & TFLOPs & GQA & TextVQA & MME & MMB & MMMU & POPE & SQA & AI2D & OCRB & VizWiz & Avg. Ret. \\
        \midrule
        \rowcolor{gray!15} LLaVA-1.5-7B & Vanilla & 100\% & 61.94 & 58.21 & 1866.15 & 64.18 & 36.11 & 85.94 & 69.46 & 55.18 & 31.50 & 54.09 & 100.0 \\
        
         & VTW (K=16) & 55\% & 54.77 & 52.17 & \textcolor{red}{1852.02} & 63.92 & \textcolor{blue!60!black}{35.67} & 86.89 & \textcolor{red}{69.71} & \textcolor{blue!60!black}{55.38} & 5.20 & 49.68 & 88.6 \\

         & ShortV (N=20) & 55\% & 60.49 & 53.63 & 1831.26 & \textcolor{blue!60!black}{64.60} & \textcolor{blue!60!black}{35.67} & 86.19 & 68.27 & 54.24 & 28.10 & 49.65 & 96.5 \\

         & VSkip (N=20)& 76\% & \textcolor{blue!60!black}{60.77} & \textcolor{red}{57.21} & 1766.56 & \textcolor{red}{64.69} & 35.56 & \textcolor{red}{87.33} & \textcolor{blue!60!black}{69.16} & 55.12 & \textcolor{red}{30.80} & \textcolor{red}{53.95} & \textcolor{red}{98.9} \\
         & VSkip\textsuperscript{+} (N=20)& 76\% & 61.63 & 57.21 & 1840.62 & 64.60 & 36.44 & 87.26 & 69.26 & 55.12 & 30.50 & 54.16 & 99.6 \\
         
         & V2Drop & 66\% & 60.51 & 55.92 & 1838.27 & 64.15 & 31.67 & 86.82 & 69.12 & 54.37 & 29.70 & 52.56 & 97.0 \\

         & APET & 66\% & 60.45 & 56.36 & 1841.29 & 64.09 & \textcolor{red}{36.11} & 86.82 & 68.93 & \textcolor{red}{55.54} & 29.90 & \textcolor{blue!60!black}{53.23} & 98.7 \\

         & Ours (N=20) & 66\% & \textcolor{red}{60.84} & \textcolor{blue!60!black}{56.40} & \textcolor{blue!60!black}{1841.68} & 64.52 & \textcolor{blue!60!black}{35.67} & \textcolor{blue!60!black}{87.26} & \textcolor{blue!60!black}{69.16} & 55.18 & \textcolor{blue!60!black}{30.20} & 52.75 & \textcolor{blue!60!black}{98.8} \\
        \midrule

        \rowcolor{gray!15} Qwen2.5-VL-7B & Vanilla & 100\% & 60.40 & 77.77 & 2515.82 & 83.25 & 50.00 & 87.62 & 87.51 & 82.42 & 84.10 & 70.81 & 100.0 \\
        
         & VTW (K=16) & 50\% & 46.01 & 65.43 & 1659.20 & 68.21 & 43.56 & 70.60 & 80.32 & 63.12 & 42.39 & 52.86 & 76.9 \\
         & ShortV (N=20) & 50\% & 49.47 & 71.93 & 1772.94 & 72.25 & 41.89 & 58.91 & 78.38 & 65.06 & 45.50 & 36.36 & 75.7 \\
         & VSkip (N=16) & 67\% & 57.26 & 63.58 & 2039.69 & 80.15 & 47.00 & 84.38 & 85.72 & 78.04 & \textcolor{blue!60!black}{69.40} & 63.85 & 91.0 \\
         & V2Drop & 55\% & 58.87 & \textcolor{red}{72.26} & 2253.59 & 80.29 & \textcolor{blue!60!black}{49.89} & \textcolor{red}{87.34} & \textcolor{blue!60!black}{87.06} & \textcolor{red}{81.54} & 56.40 & \textcolor{blue!60!black}{68.68} & 93.8 \\
         & APET & 55\% & \textcolor{red}{59.38} & 70.90 & \textcolor{red}{2263.83} & \textcolor{red}{81.70} & \textcolor{red}{50.11} & 86.70 & \textcolor{red}{87.51} & \textcolor{red}{81.54} & 60.00 & \textcolor{red}{69.02} & \textcolor{blue!60!black}{94.5} \\
         
         & Ours (N=20) & 55\% & \textcolor{blue!60!black}{59.29} & \textcolor{blue!60!black}{72.02} & \textcolor{blue!60!black}{2263.38} & \textcolor{blue!60!black}{80.67} & 46.89 & \textcolor{blue!60!black}{87.32} & 85.52 & \textcolor{blue!60!black}{79.95} & \textcolor{red}{76.90} & 65.48 & \textcolor{red}{95.0} \\
        \midrule
        \rowcolor{gray!15} Qwen3-VL-8B & Vanilla & 100\% & 61.60 & 80.07 & 2390.38 & 84.79 & 51.33 & 89.13 & 94.40 & 83.78 & 82.80 & 69.37 & 100.0 \\
        
         & VTW (K=16) & 58\% & 43.72 & 51.16 & 1808.02 & 67.53 & 47.33 & 73.88 & 81.11 & 69.88 & 37.00 & 54.81 & 75.8 \\
         & ShortV (N=20) & 58\% & 60.59 & 76.08 & 2263.21 & 81.62 & 52.22 & 88.97 & \textcolor{blue!60!black}{92.61} & 81.77 & 73.90 & 69.61 & 97.1 \\
         & VSkip (N=20) & 76\% & 60.67 & 75.98 & 2226.98 & 81.87 & 51.33 & \textcolor{blue!60!black}{89.16} & 92.46 & 80.60 & 72.20 & 68.49 & 96.3 \\
         & V2Drop & 66\% & 60.06 & 76.44 & 2290.27 & \textcolor{blue!60!black}{82.56} & 52.44 & 88.34 & 91.72 & 81.64 & 63.80 & \textcolor{blue!60!black}{69.78} & 96.0 \\
         & APET & 66\% & \textcolor{red}{60.96} & \textcolor{blue!60!black}{78.36} & \textcolor{blue!60!black}{2332.50} & 82.47 & \textcolor{blue!60!black}{52.53} & 88.54 & 92.56 & \textcolor{blue!60!black}{82.16} & \textcolor{blue!60!black}{75.80} & 68.82 & \textcolor{blue!60!black}{98.0} \\
         
         & Ours (N=20) & 66\% & \textcolor{blue!60!black}{60.70} & \textcolor{red}{78.45} & \textcolor{red}{2336.01} & \textcolor{red}{83.08} & \textcolor{red}{52.89} & \textcolor{red}{89.43} & \textcolor{red}{94.35} & \textcolor{red}{82.55} & \textcolor{red}{80.40} & \textcolor{red}{71.70} & \textcolor{red}{99.5} \\
        \bottomrule
    \end{tabular}}
    \vspace{-0.8em}
\end{table*}

\begin{table*}[t]
    \centering
    \caption{
    Sensitivity to the operator-skipping budget across the three evaluated MLLMs. 
    Avg. Ret. is the mean relative to the vanilla model.
    }
    \label{tab:budget}
    \resizebox{\textwidth}{!}{%
    \begin{tabular}{lccccccccccc}
        \toprule
        Budget & GQA & TextVQA & MME & MMB & MMMU & POPE & SQA & AI2D & OCRB & VizWiz & Avg. Ret. \\
        \midrule
        \rowcolor{gray!15}\multicolumn{12}{c}{\textit{LLaVA-1.5-7B (32 Layers)}} \\
        0 & \textbf{61.94} & \textbf{58.21} & 1866.15 & 64.18 & \textbf{36.11} & 85.94 & 69.46 & 55.18 & \textbf{31.50} & \textbf{54.09} & 100.0 \\
        8 & 61.14 & 57.18 & \textbf{1878.10} & 64.18 & 35.67 & 86.43 & \textbf{69.61} & \textbf{55.51} & 31.40 & 53.23 & 99.6 \\
        20 & 60.84 & 56.40 & 1841.68 & \textbf{64.52} & 35.67 & \textbf{87.26} & 69.16 & 55.18 & 30.20 & 52.75 & 98.8 \\
        32 & 57.89 & 53.80 & 1728.08 & 62.03 & 35.67 & 85.01 & 69.21 & 52.46 & 25.10 & 52.76 & 94.5 \\
        \midrule
        
        \rowcolor{gray!15}\multicolumn{12}{c}{\textit{Qwen2.5-VL-7B (28 Layers)}} \\
        0 & \textbf{60.40} & \textbf{77.77} & \textbf{2515.82} & \textbf{83.25} & \textbf{50.00} & \textbf{87.62} & \textbf{87.51} & 82.42 & \textbf{84.10} & \textbf{70.81} & 100.0 \\
        8 & 59.85 & 76.62 & 2346.15 & 82.56 & 49.11 & 87.29 & 87.41 & \textbf{82.58} & 81.80 & 68.25 & 98.2 \\
        20 & 59.29 & 72.02 & 2263.38 & 80.67 & 46.89 & 87.32 & 85.52 & 79.95 & 76.90 & 65.48 & 95.0 \\
        28 & 57.27 & 54.75 & 1884.70 & 69.07 & 39.33 & 86.29 & 77.94 & 62.08 & 42.30 & 54.81 & 79.2 \\
        \midrule
        \rowcolor{gray!15}\multicolumn{12}{c}{\textit{Qwen3-VL-8B (36 Layers)}} \\
        0 & \textbf{61.60} & \textbf{80.07} & \textbf{2390.38} & \textbf{84.79} & 51.33 & 89.13 & 94.40 & \textbf{83.78} & \textbf{82.80} & 69.37 & 100.0 \\
        12 & 60.98 & 79.93 & 2352.81 & 84.62 & 52.44 & 88.83 & \textbf{94.70} & 83.65 & 81.90 & 70.78 & 100.0 \\
        20 & 60.70 & 78.45 & 2336.01 & 83.08 & \textbf{52.89} & \textbf{89.43} & 94.35 & 82.55 & 80.40 & \textbf{71.70} & 99.5 \\
        28 & 59.97 & 75.58 & 2158.87 & 80.93 & 51.00 & 88.42 & 85.72 & 76.39 & 71.20 & 66.86 & 94.0 \\
        36 & 58.51 & 69.91 & 1707.86 & 75.34 & 48.22 & 87.42 & 81.85 & 72.38 & 63.80 & 63.77 & 87.7 \\
        \bottomrule
    \end{tabular}%
    }
    \vspace{-0.8em}
\end{table*}

\subsection{Complexity Analysis}
\label{sec:complexity}

Let $N_v$ be the number of visual tokens, $d$ the hidden dimension, and $d_{\mathrm{ff}}$ the FFN intermediate dimension. Since our method modifies only visual rows and keeps text-token computation unchanged, the computational reduction comes from skipping visual-row attention, visual-row FFN, or both.

\begingroup
\setlength{\abovedisplayskip}{0.35em}
\setlength{\belowdisplayskip}{0.35em}
\begin{equation}
\begin{aligned}
\Delta \mathcal{C}_{\mathcal A}
&= \Theta\left(N_v d^2 + M_{\mathcal V} d\right), \\
\Delta \mathcal{C}_{\mathcal F}
&= \Theta\left(\kappa N_v d d_{\mathrm{ff}}\right).
\end{aligned}
\end{equation}
\endgroup
where $\kappa=3$ for gated FFNs used in LLaMA- and Qwen-style backbones, and $\kappa=2$ for standard two-layer FFNs.
\begingroup
\setlength{\abovedisplayskip}{0.35em}
\setlength{\belowdisplayskip}{0.35em}
\begin{equation}
\begin{aligned}
\Delta \mathcal{C}_{\mathrm{skip}}
&= \sum_{l\in \mathcal{L}_{\mathrm{FFN}}}
\Delta \mathcal{C}_{\mathcal A}^{(l)} \\
&\quad + \sum_{l\in \mathcal{L}_{\mathrm{Attn}}}
\Delta \mathcal{C}_{\mathcal F}^{(l)} \\
&\quad + \sum_{l\in \mathcal{L}_{\mathrm{Freeze}}}
\left(
\Delta \mathcal{C}_{\mathcal A}^{(l)}
+
\Delta \mathcal{C}_{\mathcal F}^{(l)}
\right).
\end{aligned}
\end{equation}
\endgroup
Here, \textsc{FFN-only} skips visual attention, \textsc{Attention-only} skips visual FFN, and \textsc{Freeze-update} skips both operators. In practice, wall-clock speedup also depends on kernel implementation, memory movement, and hardware utilization; therefore, we report both theoretical FLOPs and measured first-token latency.

\section{Experiments}

\subsection{Experimental Setup}
\label{subsec:exp_setup}

\paragraph{Models.}
We evaluate our method on three representative open-source MLLMs, LLaVA-1.5-7B~\citep{llava15}, Qwen2.5-VL-7B~\citep{qwen25vl}, and Qwen3-VL-8B~\citep{qwen3vl}. LLaVA-1.5-7B uses a fixed-resolution visual encoder and represents each image with 576 visual tokens, while Qwen2.5-VL-7B and Qwen3-VL-8B adopt more recent Qwen-VL architectures with stronger visual encoders and dynamic-resolution image processing. This evaluation tests whether operator-level visual redundancy generalizes across different MLLM designs. All evaluations are conducted with \texttt{lmms-eval} under the official benchmark protocols.

\paragraph{Benchmarks.}
We evaluate on 10 multimodal benchmarks covering general VQA, text-rich understanding, hallucination sensitivity, multimodal reasoning, and scientific QA, GQA~\citep{gqa}, TextVQA~\citep{textvqa}, MME~\citep{mme}, MMBench~\citep{mmbench}, MMMU~\citep{mmmu}, POPE~\citep{pope}, ScienceQA~\citep{scienceqa}, AI2D~\citep{ai2d}, OCRBench~\citep{ocrbench}, and VizWiz~\citep{vizwiz}.

\paragraph{Baselines.}
We compare with training-free acceleration baselines including layer-wise visual skipping methods VTW~\citep{vtw}, ShortV~\citep{shortv}, and V-Skip~\citep{vskip}, as well as token-reduction methods V2Drop~\citep{v2drop} and APET~\citep{apet}. Unlike these coarse-grained baselines, our method preserves all visual tokens and selectively skips attention, FFN, or both according to layer-wise operator dominance.

\paragraph{Implementation details.}
For each model, we estimate layer-wise visual operator risks using a small calibration set, sampling $7$ examples from each of GQA, POPE, TextVQA, MME, MMMU, and ScienceQA. These examples are used only for estimating $\mathcal{R}{\mathcal A}(l)$, $\mathcal{R}{\mathcal F}(l)$, and $\mathcal{R}_{\mathcal{AF}}(l)$. Based on the risks, each selected layer is assigned one of three policies, \textsc{Attention-only}, \textsc{FFN-only}, or \textsc{Freeze-update}. The resulting layer-wise policy is fixed for all samples of the same model. We report relative FLOPs ratio, and first-token latency measured on NVIDIA RTX A800 GPUs. 

\subsection{Main Results}
\label{subsec:main_results}

Table~\ref{tab:main_results} compares our method with representative layer-wise skipping and token-pruning baselines across three MLLM architectures and ten benchmarks. Our method achieves a strong accuracy--efficiency trade-off by preserving the full visual-token sequence while selectively skipping redundant operators, leading to the highest average performance retention across the evaluated backbones.

This robustness comes from keeping the dominant visual operator in each selected layer, which helps preserve fine-grained and text-rich evidence while removing answer-silent computation. Following the V-Skip paper, VSkip\textsuperscript{+} uses dataset-specific optimal settings, whereas the other methods use one configuration for all benchmarks on each model. Our single per-model policy remains competitive or superior under this comparison setting.

\subsection{Efficiency Comparison on TextVQA}
\label{subsec:textvqa_efficiency}

We further compare efficiency on TextVQA against ShortV, a competitive layer-wise visual-computation baseline. As shown in Table~\ref{tab:efficiency}, ShortV reduces more FLOPs but causes larger accuracy degradation, whereas our method achieves comparable prefill efficiency and first-token latency while retaining stronger answer quality. This shows that operator-level skipping provides a better accuracy--efficiency balance than freezing full visual-token updates.
\begin{table}[htbp]
    \centering
    \caption{
    Efficiency comparison with ShortV on TextVQA. 
    We report theoretical FLOPs, relative FLOPs ratio, first-token latency, and average score.
    }
    \label{tab:efficiency}
    \resizebox{\columnwidth}{!}{%
    \begin{tabular}{lcccc}
        \toprule
        Model & FLOPs & Prefill & Latency & Avg. \\
        \midrule
        LLaVA-1.5-7B & 9.057 T & 59.1 ms & 120.4 ms & 100.0 \\
        ShortV & 5.039 T & 45.0 ms & 108.2 ms & 96.5 \\
        Ours & 6.475 T & 49.3 ms & 110.4 ms & 98.8 \\
        \midrule
        Qwen3-VL-8B & 11.38 T & 95.7 ms & 465.1 ms & 100.0 \\
        ShortV & 5.79 T & 45.8 ms & 310.6 ms & 97.1 \\
        Ours & 7.23 T & 52.9 ms & 344.8 ms & 99.5 \\
        \bottomrule
    \end{tabular}%
    }
\end{table}

\begin{figure*}[t]
    \centering
    \includegraphics[width=\textwidth]{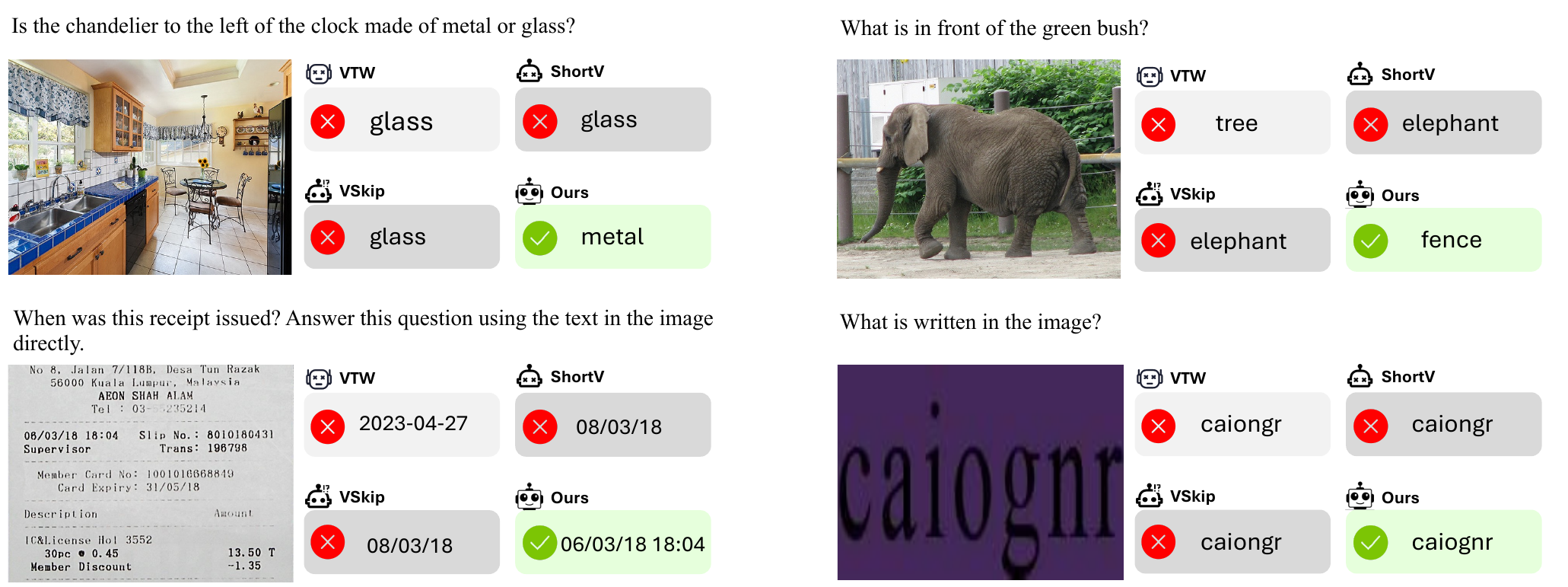}
    \caption{Qualitative visualization of model responses under operator-aware visual-token skipping.}
    \label{fig:qualitative_examples}
    \vspace{-0.8em}
\end{figure*}

Figure~\ref{fig:qualitative_examples} visualizes model responses under operator-aware visual-token skipping. Our method still produces accurate answers on both reasoning and recognition examples, suggesting that it preserves the operators responsible for key layer-wise visual reasoning functions and largely maintains the model's answer quality.

\subsection{Effect of Operator-Skipping Budget}
\label{subsec:budget}

We study how the operator-skipping budget affects performance across the three evaluated MLLMs. As shown in Table~\ref{tab:budget}, moderate budgets simplify redundant visual computation through \textsc{Attention-only}, \textsc{FFN-only}, or \textsc{Freeze-update} while preserving most benchmark scores, whereas overly aggressive budgets eventually degrade performance on fine-grained or text-rich tasks. This consistent trend across LLaVA-1.5-7B, Qwen2.5-VL-7B, and Qwen3-VL-8B suggests that operator-level redundancy is broadly present, but the best budget remains model-dependent and should be treated as an efficiency--performance knob.

\subsection{Ablation Study}
\label{subsec:ablation_qwen3_28}

To isolate the contribution of operator-aware policy assignment, we conduct a fixed-budget ablation on Qwen3-VL-8B with budget $B=28$. All variants simplify the same number of layers, but use different operator actions, \textsc{Attn-Skip} always skips visual attention, \textsc{FFN-Skip} always skips visual FFN, and \textsc{Frozen} skips both operators. As shown in Table~\ref{tab:ablation_qwen3_28}, fixed rules are not uniformly reliable, while our layer-wise risk-based assignment better balances efficiency and accuracy by matching the skipped operator to each layer's dominant visual operator.

\begin{table}[htbp]
\centering
\caption{
Ablation on Qwen3-VL-8B under budget $B=28$. 
All variants simplify the same number of layers.
}
\label{tab:ablation_qwen3_28}
\resizebox{\columnwidth}{!}{
\begin{tabular}{lccccc}
\toprule
Method & GQA & POPE & MME & OCRB & VizWiz \\
\midrule
\rowcolor{gray!15} Vanilla & 61.6 & 89.1 & 2390 & 82.8 & 69.4 \\
\textsc{Attn-Skip} & 53.6 & 78.6 & 1506 & 11.2 & 58.8 \\
\textsc{FFN-Skip} & 59.6 & \textbf{89.3} & 2118 & 61.7 & 66.2 \\
\textsc{Frozen} & 26.8 & 57.1 & 975 & 20.0 & 52.3 \\
Ours & \textbf{60.0} & 89.2 & \textbf{2159} & \textbf{76.4} & \textbf{69.1} \\
\bottomrule
\end{tabular}}

\end{table}

\subsection{Combination with Token Pruning}
\label{subsec:token_pruning_combination}

Finally, we study whether operator-aware skipping is complementary to token pruning. Token pruning reduces the visual sequence length, while our method removes redundant operator computation on the remaining tokens, so the two strategies target different sources of redundancy. As shown in Table~\ref{tab:token_pruning_combination}, applying our method on top of VisionZip~\citep{visionzip} or V2Drop~\citep{v2drop} further reduces computation while preserving answer quality better than more aggressive pruning. This suggests that, especially in the low-token regime, retaining more visual evidence and skipping redundant operators is preferable to further shortening the visual sequence.

\begin{table}[htbp]
\centering
\caption{
Combining operator-aware skipping with token pruning on Qwen3-VL-8B. 
}
\label{tab:token_pruning_combination}
\resizebox{\columnwidth}{!}{
\begin{tabular}{lcccccc}
\toprule
Method (Token Ret.) & TFLOPs & GQA & TextVQA & MME & MMB & POPE \\
\midrule

VisionZip ($10\%$) & 1.67 & 49.8 & 61.0 & 1717.1 & 69.0 & 79.5 \\
VisionZip ($3\%$) & \downchange{1.33}{0.34} & \downchange{42.0}{7.8} & \downchange{46.6}{14.4} & \downchange{1412.9}{304.2} & \downchange{47.9}{21.1} & \downchange{65.5}{14.0} \\
VisionZip + Ours ($10\%$) & \downchange{1.31}{0.36} & \upchange{49.9}{0.1} & \downchange{60.7}{0.3} & \upchange{1723.7}{6.6} & \downchange{66.3}{2.7} & \upchange{80.3}{0.8} \\
\midrule

V2Drop ($10\%$) & 2.59 & 54.7 & 60.4 & 2012.6 & 77.7 & 86.5 \\
V2Drop ($2\%$) & \downchange{1.99}{0.60} & \downchange{41.5}{13.2} & \downchange{44.9}{15.5} & \downchange{1482.2}{530.4} & \downchange{44.1}{33.6} & \downchange{67.5}{19.0} \\
V2Drop + Ours ($10\%$) & \downchange{1.31}{1.28} & \upchange{55.6}{0.9} & \upchange{60.9}{0.5} & \downchange{1999.2}{13.4} & \downchange{77.4}{0.3} & \downchange{86.4}{0.1} \\
\bottomrule
\end{tabular}}

\end{table}

\subsection{Qualitative Comparison}
\label{subsec:qualitative_comparison}

Figure~\ref{fig:qualitative_examples} shows that operator-aware visual-token skipping preserves accurate responses on examples requiring visual reasoning and object recognition. These qualitative results suggest that our policy maintains answer capability while removing redundant visual computation.

\section{Conclusion}

We demonstrated that visual redundancy in MLLMs is answer-observable and operator-dependent, with late visual-token updates often contributing little to answers. Our operator-aware skipping preserves all visual tokens while bypassing redundant attention, FFN, or both, reducing visual computation across multiple backbones with little performance loss, without irreversible evidence loss from pruning.


\clearpage

\bibliography{main}

\begin{thebibliography}{32}
\providecommand{\natexlab}[1]{#1}

\bibitem[{Alvar et~al.(2025)Alvar, Singh, Akbari, and Zhang}]{divprune}
Saeed~Ranjbar Alvar, Gursimran Singh, Mohammad Akbari, and Yong Zhang. 2025.
\newblock Divprune: Diversity-based visual token pruning for large multimodal models.
\newblock In \emph{Proceedings of the IEEE/CVF Conference on Computer Vision and Pattern Recognition (CVPR)}, pages 9392--9401.

\bibitem[{An et~al.(2025)An, Xie, Yang, Zhang, Zhao, Cheng, Wang, Xu, Chen, Zhu, Wu, Tan, Li, Yang, Yu, Wang, Qin, Wang, Yan, Feng, Liu, Li, and Deng}]{llavaonevision}
Xiang An, Yin Xie, Kaicheng Yang, Wenkang Zhang, Xiuwei Zhao, Zheng Cheng, Yirui Wang, Songcen Xu, Changrui Chen, Didi Zhu, Chunsheng Wu, Huajie Tan, Chunyuan Li, Jing Yang, Jie Yu, Xiyao Wang, Bin Qin, Yumeng Wang, Zizhen Yan, and 4 others. 2025.
\newblock \href {https://arxiv.org/abs/2509.23661} {Llava-onevision-1.5: Fully open framework for democratized multimodal training}.
\newblock \emph{Preprint}, arXiv:2509.23661.

\bibitem[{Bai et~al.(2025{\natexlab{a}})Bai, Cai, Chen, Chen, Chen, Cheng, Deng, Ding, Gao, Ge, Ge, Guo, Huang, Huang, Huang, Hui, Jiang, Li, Li, Li, Li, Lin, Lin, Liu, Liu, Liu, Liu, Liu, Liu, Lu, Luo, Lv, Men, Meng, Ren, Ren, Song, Sun, Tang, Tu, Wan, Wang, Wang, Wang, Wang, Xie, Xu, Xu, Xu, Yang, Yang, Yang, Yang, Yu, Zhang, Zhang, Zhang, Zheng, Zhong, Zhou, Zhou, Zhou, Zhu, and Zhu}]{qwen3vl}
Shuai Bai, Yuxuan Cai, Ruizhe Chen, Keqin Chen, Xionghui Chen, Zesen Cheng, Lianghao Deng, Wei Ding, Chang Gao, Chunjiang Ge, Wenbin Ge, Zhifang Guo, Qidong Huang, Jie Huang, Fei Huang, Binyuan Hui, Shutong Jiang, Zhaohai Li, Mingsheng Li, and 45 others. 2025{\natexlab{a}}.
\newblock \href {https://arxiv.org/abs/2511.21631} {Qwen3-vl technical report}.
\newblock \emph{Preprint}, arXiv:2511.21631.

\bibitem[{Bai et~al.(2025{\natexlab{b}})Bai, Chen, Liu, Wang, Ge, Song, Dang, Wang, Wang, Tang, Zhong, Zhu, Yang, Li, Wan, Wang, Ding, Fu, Xu, Ye, Zhang, Xie, Cheng, Zhang, Yang, Xu, and Lin}]{qwen25vl}
Shuai Bai, Keqin Chen, Xuejing Liu, Jialin Wang, Wenbin Ge, Sibo Song, Kai Dang, Peng Wang, Shijie Wang, Jun Tang, Humen Zhong, Yuanzhi Zhu, Mingkun Yang, Zhaohai Li, Jianqiang Wan, Pengfei Wang, Wei Ding, Zheren Fu, Yiheng Xu, and 8 others. 2025{\natexlab{b}}.
\newblock \href {https://arxiv.org/abs/2502.13923} {Qwen2.5-vl technical report}.
\newblock \emph{Preprint}, arXiv:2502.13923.

\bibitem[{Chen et~al.(2026)Chen, Liu, Wen, Wang, Huang, and Chen}]{v2drop}
Junjie Chen, Xuyang Liu, Zichen Wen, Yiyu Wang, Siteng Huang, and Honggang Chen. 2026.
\newblock Variation-aware vision token dropping for faster large vision-language models.
\newblock In \emph{Proceedings of the IEEE/CVF Conference on Computer Vision and Pattern Recognition (CVPR)}, pages 3489--3499.

\bibitem[{Chen et~al.(2025)Chen, Zhao, Liu, Bai, Lin, Zhou, and Chang}]{fastv}
Liang Chen, Haozhe Zhao, Tianyu Liu, Shuai Bai, Junyang Lin, Chang Zhou, and Baobao Chang. 2025.
\newblock An image is worth 1/2 tokens after layer 2: Plug-and-play inference acceleration for large vision-language models.
\newblock In \emph{Computer Vision -- ECCV 2024}, pages 19--35, Cham. Springer Nature Switzerland.

\bibitem[{Fu et~al.(2025)Fu, Chen, Shen, Qin, Zhang, Lin, Yang, Zheng, Li, Sun, Wu, Ji, Shan, and He}]{mme}
Chaoyou Fu, Peixian Chen, Yunhang Shen, Yulei Qin, Mengdan Zhang, Xu~Lin, Jinrui Yang, Xiawu Zheng, Ke~Li, Xing Sun, Yunsheng Wu, Rongrong Ji, Caifeng Shan, and Ran He. 2025.
\newblock \href {https://proceedings.neurips.cc/paper_files/paper/2025/file/d79a27cf2772fe00be7f341efc0eb517-Paper-Datasets_and_Benchmarks_Track.pdf} {Mme: A comprehensive evaluation benchmark for multimodal large language models}.
\newblock In \emph{Advances in Neural Information Processing Systems}, volume~38. Curran Associates, Inc.

\bibitem[{Gurari et~al.(2018)Gurari, Li, Stangl, Guo, Lin, Grauman, Luo, and Bigham}]{vizwiz}
Danna Gurari, Qing Li, Abigale~J. Stangl, Anhong Guo, Chi Lin, Kristen Grauman, Jiebo Luo, and Jeffrey~P. Bigham. 2018.
\newblock Vizwiz grand challenge: Answering visual questions from blind people.
\newblock In \emph{Proceedings of the IEEE Conference on Computer Vision and Pattern Recognition (CVPR)}.

\bibitem[{Huang et~al.(2026)Huang, Ma, Shao, Guo, Yu, Cui, and Tian}]{nuwa}
Yihong Huang, Fei Ma, Yihua Shao, Jingcai Guo, Zitong Yu, Laizhong Cui, and Qi~Tian. 2026.
\newblock \href {https://arxiv.org/abs/2602.02951} {N\"uwa: Mending the spatial integrity torn by vlm token pruning}.
\newblock \emph{Preprint}, arXiv:2602.02951.

\bibitem[{Hudson and Manning(2019)}]{gqa}
Drew~A. Hudson and Christopher~D. Manning. 2019.
\newblock Gqa: A new dataset for real-world visual reasoning and compositional question answering.
\newblock In \emph{Proceedings of the IEEE/CVF Conference on Computer Vision and Pattern Recognition (CVPR)}.

\bibitem[{Kembhavi et~al.(2016)Kembhavi, Salvato, Kolve, Seo, Hajishirzi, and Farhadi}]{ai2d}
Aniruddha Kembhavi, Mike Salvato, Eric Kolve, Minjoon Seo, Hannaneh Hajishirzi, and Ali Farhadi. 2016.
\newblock A diagram is worth a dozen images.
\newblock In \emph{Computer Vision -- ECCV 2016}, pages 235--251, Cham. Springer International Publishing.

\bibitem[{Kim et~al.(2026)Kim, Kim, Kim, Lee, and Park}]{tokenpruningfails}
Jiwan Kim, Kibum Kim, Wonjoong Kim, Byung-Kwan Lee, and Chanyoung Park. 2026.
\newblock \href {https://arxiv.org/abs/2604.12358} {Why and when visual token pruning fails? a study on relevant visual information shift in mllms decoding}.
\newblock \emph{Preprint}, arXiv:2604.12358.

\bibitem[{Li et~al.(2023{\natexlab{a}})Li, Li, Savarese, and Hoi}]{blip2}
Junnan Li, Dongxu Li, Silvio Savarese, and Steven Hoi. 2023{\natexlab{a}}.
\newblock \href {https://proceedings.mlr.press/v202/li23q.html} {{BLIP}-2: Bootstrapping language-image pre-training with frozen image encoders and large language models}.
\newblock In \emph{Proceedings of the 40th International Conference on Machine Learning}, volume 202 of \emph{Proceedings of Machine Learning Research}, pages 19730--19742. PMLR.

\bibitem[{Li et~al.(2023{\natexlab{b}})Li, Du, Zhou, Wang, Zhao, and Wen}]{pope}
Yifan Li, Yifan Du, Kun Zhou, Jinpeng Wang, Xin Zhao, and Ji-Rong Wen. 2023{\natexlab{b}}.
\newblock \href {https://doi.org/10.18653/v1/2023.emnlp-main.20} {Evaluating object hallucination in large vision-language models}.
\newblock In \emph{Proceedings of the 2023 Conference on Empirical Methods in Natural Language Processing}, pages 292--305, Singapore. Association for Computational Linguistics.

\bibitem[{Lin et~al.(2025)Lin, Lin, Lin, and Ji}]{vtw}
Zhihang Lin, Mingbao Lin, Luxi Lin, and Rongrong Ji. 2025.
\newblock \href {https://doi.org/10.1609/aaai.v39i5.32567} {Boosting multimodal large language models with visual tokens withdrawal for rapid inference}.
\newblock \emph{Proceedings of the AAAI Conference on Artificial Intelligence}, 39(5):5334--5342.

\bibitem[{Liu et~al.(2024{\natexlab{a}})Liu, Li, Li, and Lee}]{llava15}
Haotian Liu, Chunyuan Li, Yuheng Li, and Yong~Jae Lee. 2024{\natexlab{a}}.
\newblock Improved baselines with visual instruction tuning.
\newblock In \emph{Proceedings of the IEEE/CVF Conference on Computer Vision and Pattern Recognition (CVPR)}, pages 26296--26306.

\bibitem[{Liu et~al.(2024{\natexlab{b}})Liu, Li, Li, Li, Zhang, Shen, and Lee}]{llava-next}
Haotian Liu, Chunyuan Li, Yuheng Li, Bo~Li, Yuanhan Zhang, Sheng Shen, and Yong~Jae Lee. 2024{\natexlab{b}}.
\newblock \href {https://llava-vl.github.io/blog/2024-01-30-llava-next/} {Llava-next: Improved reasoning, ocr, and world knowledge}.

\bibitem[{Liu et~al.(2025)Liu, Duan, Zhang, Li, Zhang, Zhao, Yuan, Wang, He, Liu, Chen, and Lin}]{mmbench}
Yuan Liu, Haodong Duan, Yuanhan Zhang, Bo~Li, Songyang Zhang, Wangbo Zhao, Yike Yuan, Jiaqi Wang, Conghui He, Ziwei Liu, Kai Chen, and Dahua Lin. 2025.
\newblock Mmbench: Is your multi-modal model an all-around player?
\newblock In \emph{Computer Vision -- ECCV 2024}, pages 216--233, Cham. Springer Nature Switzerland.

\bibitem[{Liu et~al.(2024{\natexlab{c}})Liu, Li, Huang, Yang, Yu, Li, Yin, Liu, Jin, and Bai}]{ocrbench}
Yuliang Liu, Zhang Li, Mingxin Huang, Biao Yang, Wenwen Yu, Chunyuan Li, Xu-Cheng Yin, Cheng-Lin Liu, Lianwen Jin, and Xiang Bai. 2024{\natexlab{c}}.
\newblock \href {https://doi.org/10.1007/s11432-024-4235-6} {{OCRBench}: On the hidden mystery of {OCR} in large multimodal models}.
\newblock \emph{Science China Information Sciences}, 67(12):220102.

\bibitem[{Lu et~al.(2022)Lu, Mishra, Xia, Qiu, Chang, Zhu, Tafjord, Clark, and Kalyan}]{scienceqa}
Pan Lu, Swaroop Mishra, Tanglin Xia, Liang Qiu, Kai-Wei Chang, Song-Chun Zhu, Oyvind Tafjord, Peter Clark, and Ashwin Kalyan. 2022.
\newblock \href {https://proceedings.neurips.cc/paper_files/paper/2022/file/11332b6b6cf4485b84afadb1352d3a9a-Paper-Conference.pdf} {Learn to explain: Multimodal reasoning via thought chains for science question answering}.
\newblock In \emph{Advances in Neural Information Processing Systems}, volume~35, pages 2507--2521. Curran Associates, Inc.

\bibitem[{Ma et~al.(2026{\natexlab{a}})Ma, Qiu, Ji, Sun, and Ji}]{vskip}
Jie Ma, Zhike Qiu, Jiayi Ji, Xiaoshuai Sun, and Rongrong Ji. 2026{\natexlab{a}}.
\newblock \href {https://arxiv.org/abs/2606.08511} {Look less, reason more: Block-wise attention skipping for efficient multimodal llms}.
\newblock \emph{Preprint}, arXiv:2606.08511.

\bibitem[{Ma et~al.(2026{\natexlab{b}})Ma, Zhang, Wang, Song, Chen, and Zheng}]{apet}
Qiankun Ma, Ziyao Zhang, Haofei Wang, Zhen Song, Jie Chen, and Hairong Zheng. 2026{\natexlab{b}}.
\newblock Apet: Approximation-error guided token compression for efficient vlms.
\newblock In \emph{Proceedings of the IEEE/CVF Conference on Computer Vision and Pattern Recognition (CVPR)}, pages 26306--26316.

\bibitem[{Shang et~al.(2025)Shang, Cai, Xu, Lee, and Yan}]{llava_prumerge}
Yuzhang Shang, Mu~Cai, Bingxin Xu, Yong~Jae Lee, and Yan Yan. 2025.
\newblock Llava-prumerge: Adaptive token reduction for efficient large multimodal models.
\newblock In \emph{Proceedings of the IEEE/CVF International Conference on Computer Vision (ICCV)}, pages 22857--22867.

\bibitem[{Singh et~al.(2019)Singh, Natarajan, Shah, Jiang, Chen, Batra, Parikh, and Rohrbach}]{textvqa}
Amanpreet Singh, Vivek Natarajan, Meet Shah, Yu~Jiang, Xinlei Chen, Dhruv Batra, Devi Parikh, and Marcus Rohrbach. 2019.
\newblock Towards vqa models that can read.
\newblock In \emph{Proceedings of the IEEE/CVF Conference on Computer Vision and Pattern Recognition (CVPR)}.

\bibitem[{Wang et~al.(2026)Wang, Wu, Ni, Yang, Liu, Yang, Wen, He, Tang, Liu, and Zhou}]{randomtoken}
Yahong Wang, Juncheng Wu, Zhangkai Ni, Longzhen Yang, Yihang Liu, Chengmei Yang, Ying Wen, Lianghua He, Xianfeng Tang, Hui Liu, and Yuyin Zhou. 2026.
\newblock When token pruning is worse than random: Understanding visual token information in vllms.
\newblock In \emph{Proceedings of the IEEE/CVF Conference on Computer Vision and Pattern Recognition (CVPR)}, pages 31910--31919.

\bibitem[{Wen et~al.(2025)Wen, Gao, Li, He, and Zhang}]{solvingright}
Zichen Wen, Yifeng Gao, Weijia Li, Conghui He, and Linfeng Zhang. 2025.
\newblock \href {https://doi.org/10.18653/v1/2025.findings-acl.802} {Token pruning in multimodal large language models: Are we solving the right problem?}
\newblock In \emph{Findings of the Association for Computational Linguistics: ACL 2025}, pages 15537--15549, Vienna, Austria. Association for Computational Linguistics.

\bibitem[{Xi et~al.(2026)Xi, Tian, Yang, Yi, Lin, Hao, Wang, and Wang}]{lost_in_attention}
Gongli Xi, Ye~Tian, Mengyu Yang, Huahui Yi, Liang Lin, Xiaoshuai Hao, Kun Wang, and Wendong Wang. 2026.
\newblock \href {https://arxiv.org/abs/2605.05668} {Large vision-language models get lost in attention}.
\newblock \emph{Preprint}, arXiv:2605.05668.

\bibitem[{Xing et~al.(2025)Xing, Huang, Dong, Lu, Zhang, Zang, Cao, He, Wang, Wu, and Lin}]{pdrop}
Long Xing, Qidong Huang, Xiaoyi Dong, Jiajie Lu, Pan Zhang, Yuhang Zang, Yuhang Cao, Conghui He, Jiaqi Wang, Feng Wu, and Dahua Lin. 2025.
\newblock \href {https://arxiv.org/abs/2410.17247} {Pyramiddrop: Accelerating your large vision-language models via pyramid visual redundancy reduction}.
\newblock \emph{Preprint}, arXiv:2410.17247.

\bibitem[{Yang et~al.(2025)Yang, Chen, Tian, Wang, Li, Yu, and Jia}]{visionzip}
Senqiao Yang, Yukang Chen, Zhuotao Tian, Chengyao Wang, Jingyao Li, Bei Yu, and Jiaya Jia. 2025.
\newblock Visionzip: Longer is better but not necessary in vision language models.
\newblock In \emph{Proceedings of the IEEE/CVF Conference on Computer Vision and Pattern Recognition (CVPR)}, pages 19792--19802.

\bibitem[{Yuan et~al.(2025)Yuan, Zhang, Liu, Chen, Lu, Lin, Zheng, Han, and Sun}]{shortv}
Qianhao Yuan, Qingyu Zhang, Yanjiang Liu, Jiawei Chen, Yaojie Lu, Hongyu Lin, Jia Zheng, Xianpei Han, and Le~Sun. 2025.
\newblock Shortv: Efficient multimodal large language models by freezing visual tokens in ineffective layers.
\newblock In \emph{Proceedings of the IEEE/CVF International Conference on Computer Vision (ICCV)}, pages 329--339.

\bibitem[{Yue et~al.(2024)Yue, Ni, Zhang, Zheng, Liu, Zhang, Stevens, Jiang, Ren, Sun, Wei, Yu, Yuan, Sun, Yin, Zheng, Yang, Liu, Huang, Sun, Su, and Chen}]{mmmu}
Xiang Yue, Yuansheng Ni, Kai Zhang, Tianyu Zheng, Ruoqi Liu, Ge~Zhang, Samuel Stevens, Dongfu Jiang, Weiming Ren, Yuxuan Sun, Cong Wei, Botao Yu, Ruibin Yuan, Renliang Sun, Ming Yin, Boyuan Zheng, Zhenzhu Yang, Yibo Liu, Wenhao Huang, and 3 others. 2024.
\newblock Mmmu: A massive multi-discipline multimodal understanding and reasoning benchmark for expert agi.
\newblock In \emph{Proceedings of the IEEE/CVF Conference on Computer Vision and Pattern Recognition (CVPR)}, pages 9556--9567.

\bibitem[{Zhang et~al.(2025)Zhang, Liu, Li, Lu, Zhang, Pan, She, and Zhang}]{cdpruner}
Qizhe Zhang, Mengzhen Liu, Lichen Li, Ming Lu, Yuan Zhang, Junwen Pan, Qi~She, and Shanghang Zhang. 2025.
\newblock \href {https://proceedings.neurips.cc/paper_files/paper/2025/file/2433fec2144ccf5fea1c9c5ebdbc3924-Paper-Conference.pdf} {Beyond attention or similarity: Maximizing conditional diversity for token pruning in mllms}.
\newblock In \emph{Advances in Neural Information Processing Systems}, volume~38, pages 25438--25468. Curran Associates, Inc.

\end{thebibliography}

\clearpage

\appendix

\section{Related Work}
\label{sec:related_work}

\paragraph{Layer-wise visual-computation skipping.}
As high-resolution inputs, image tiling, and dense visual encoders increase the length of visual prefixes, the LLM prefill stage has become a major bottleneck for MLLM inference~\citep{fastv,shortv,vtw,vskip}. One line of work reduces this cost by modifying how visual tokens are processed across Transformer layers. VTW withdraws visual tokens from later layers to avoid repeatedly propagating visual states through the full LLM backbone~\citep{vtw}, while ShortV freezes visual-token states in layers identified as visually ineffective~\citep{shortv}. V-Skip studies block-wise attention skipping and shows that visual attention computation can be redundant in selected layers~\citep{vskip}. Related structural acceleration methods also exploit layer-wise or block-wise redundancy to reduce visual computation without directly deleting all visual inputs~\citep{fastv,vskip}. Together, these studies show that visual computation can be redundant in a depth-dependent manner, but they differ in the granularity at which redundancy is modeled. VTW and ShortV primarily act on full visual-token states, whereas V-Skip focuses on attention computation. Our work is complementary, instead of treating a layer or a single operator family as the default unit of analysis, we compare the answer-observable risks of visual-row attention and visual-row FFN within each layer. This operator-level view allows the policy to preserve the operator that remains useful at a given depth while skipping the redundant one, or to freeze the visual update when both operators are answer-silent.

\paragraph{Visual-token pruning and compression.}
Another major direction accelerates MLLM inference by reducing the number of visual tokens. Early and recent methods prune, merge, or select visual tokens according to attention scores, token similarity, spatial redundancy, diversity, or output sensitivity~\citep{fastv,llava_prumerge,visionzip,pdrop,cdpruner,divprune,apet,v2drop}. These approaches directly shorten the visual sequence and can reduce both attention and FFN costs in subsequent layers. However, token reduction is inherently destructive, once visual tokens are removed, merged, or prematurely withdrawn, the corresponding spatial evidence cannot be recovered by later textual reasoning. This limitation is especially pronounced in fine-grained recognition tasks, including OCR-centric VQA, chart or diagram understanding, small-object reasoning, and visually grounded question answering, where the decisive evidence may occupy only a small image region or become relevant only after interacting with the question. Recent studies further show that aggressive token pruning can damage spatial integrity, shift relevant visual information during decoding, and even underperform random selection in certain settings~\citep{solvingright,nuwa,tokenpruningfails,randomtoken}. In contrast, our method preserves the full visual-token sequence throughout inference and reduces computation at the operator level. This avoids irreversible loss of fine-grained visual evidence while still removing answer-silent visual updates, matching the motivation in the introduction: efficient MLLM inference should preserve visual information and reduce only computation that is not observable from the answer side.

\section{Implementation Environment}
\label{app:implementation_environment}

All experiments were conducted on an internal GPU cluster equipped with NVIDIA A800-SXM4 GPUs. Each compute node contains two Intel(R) Xeon(R) Platinum 8358 CPUs at 2.60GHz, providing 128 logical CPU cores in total, approximately 1 TiB of system memory, and 8 NVIDIA A800-SXM4 GPUs with 80GB VRAM each. The NVIDIA driver version is 535.161.08. The system CUDA toolkit is CUDA 12.2, while the PyTorch binaries used in our conda environments are built with CUDA 12.1.
\begin{table}[h]
\centering
\caption{Conda environments used for different model families.}
\label{tab:env_versions}
\resizebox{\linewidth}{!}{
\begin{tabular}{l c c c c c c}
\toprule
Model family & Python & PyTorch & Torchvision & Transformers & Accelerate & lmms-eval \\
\midrule
Qwen2-VL   & 3.12.12 & 2.4.1       & 0.19.1       & 4.51.3      & 1.13.0 & 0.6.1 \\
Qwen3-VL   & 3.12.12 & 2.4.1       & 0.19.1       & 5.5.4       & 1.13.0 & 0.6.1 \\
LLaVA-1.5  & 3.12.9  & 2.4.1       & 0.19.1       & 4.37.2      & 0.21.0 & 0.6.1 \\
LLaVA-NeXT & 3.10.20 & 2.1.2+cu121 & 0.16.2+cu121 & 4.40.0.dev0 & 0.29.3 & 0.6.1 \\
\bottomrule
\end{tabular}
}
\end{table}

We use the official \texttt{lmms-eval} framework for all reported multimodal benchmark results and follow its default evaluation protocols. Since different model families rely on different dependency constraints, we maintain separate conda environments for Qwen2-VL, Qwen3-VL, LLaVA-1.5, and LLaVA-NeXT. Model checkpoints are loaded from local copies of the corresponding public Hugging Face repositories. The evaluated benchmark suite includes MME, MMBench, OCRBench, POPE, GQA, TextVQA, ScienceQA, MMMU, AI2D, and VizWiz when applicable.
\begin{table}[h]
\centering
\caption{Additional package versions used in each environment.}
\label{tab:package_versions}
\resizebox{\linewidth}{!}{
\begin{tabular}{l c c c c}
\toprule
Model family & \texttt{Qwen2.5VL} & \texttt{Qwen3VL} & \texttt{Llava1.5} & \texttt{Llava-Next} \\
\midrule
\texttt{datasets}      & 2.16.1       & 2.16.1       & 2.16.1       & 5.0.0 \\
\texttt{tokenizers}    & 0.21.4       & 0.22.2       & 0.15.1       & 0.15.2 \\
\texttt{sentencepiece} & 0.2.1        & 0.2.1        & 0.2.1        & 0.2.1 \\
\texttt{qwen-vl-utils} & 0.0.14       & 0.0.14       & --       & -- \\
\texttt{llava}         & --  & --  & 1.2.2.post1  & 1.7.0.dev0 \\
\bottomrule
\end{tabular}
}
\end{table}

\section{Computation of Answer-Observable Metrics}
\label{app:answer_observable_computation}

In Section~\ref{sec:answer_silent_updates}, we define answer-observable influence (AOI) using the exact Jacobian from visual-token states to the final prompt-token hidden state. This definition is mathematically direct, but computing the full Jacobian is prohibitively expensive for modern MLLMs. We therefore estimate AOI with randomized probes. This section derives the estimator used in our implementation.

Let $H_v^{l+1} \in \mathbf{R}^{N_v \times d}$ denote the visual-token hidden states after layer $l$, and let
\[
    \Delta V_l = H_v^{l+1} - H_v^l
\]
be the visual-token update. We write $v_l=\mathrm{vec}(\Delta V_l)$ for its flattened representation. Let $h_{\mathrm{last}}\in\mathbf{R}^{d}$ be the final prompt-token hidden state after prefill, and let
\[
    J_l =
    \frac{\partial h_{\mathrm{last}}}
    {\partial \mathrm{vec}(H_v^{l+1})}
    \in \mathbf{R}^{d \times N_vd}
\]
be the Jacobian mapping perturbations of visual-token states at layer $l+1$ to perturbations of the final prompt-token hidden state. The exact first-order change induced by the visual update is
\[
    \delta h_l = J_l v_l,
\]
and the exact answer-observable influence is
\[
    \mathrm{AOI}_l = \|\delta h_l\|_2
    =
    \|J_l v_l\|_2 .
\]
Directly forming $J_l$ is infeasible because it has size $d \times N_vd$. Instead, we estimate the norm of $J_l v_l$ through random projections.

Let $r_k \in \mathbf{R}^{d}$ be an independent random probe satisfying
\[
    \mathbf{E}[r_k r_k^\top] = I_d .
\]
In practice, $r_k$ can be sampled from a standard Gaussian distribution or a Rademacher distribution. For each probe, we define
\[
    g_{l,k}
    =
    \left\langle
    \frac{\partial (r_k^\top h_{\mathrm{last}})}
    {\partial H_v^{l+1}},
    \Delta V_l
    \right\rangle_F .
\]
This quantity can be computed with a single reverse-mode gradient call, without explicitly constructing the full Jacobian. By the chain rule,
\[
    \frac{\partial (r_k^\top h_{\mathrm{last}})}
    {\partial \mathrm{vec}(H_v^{l+1})}
    =
    J_l^\top r_k .
\]
Therefore,
\[
    g_{l,k}
    =
    \langle J_l^\top r_k, v_l\rangle
    =
    r_k^\top J_l v_l
    =
    r_k^\top \delta h_l .
\]
Thus, $g_{l,k}$ is a random projection of the exact answer-side perturbation $\delta h_l$.

The squared projection gives an unbiased estimator of the squared AOI:
\[
\begin{aligned}
    \mathbf{E}_{r_k}[g_{l,k}^2]
    &=
    \mathbf{E}_{r_k}
    \left[(r_k^\top \delta h_l)^2\right] \\
    &=
    \delta h_l^\top
    \mathbf{E}[r_k r_k^\top]
    \delta h_l \\
    &=
    \|\delta h_l\|_2^2 \\
    &=
    \mathrm{AOI}_l^2 .
\end{aligned}
\]
With $K$ independent probes, we estimate AOI as
\[
    \widehat{\mathrm{AOI}}_l
    =
    \left(
    \frac{1}{K}
    \sum_{k=1}^{K}
    g_{l,k}^2
    \right)^{1/2}.
\]
This estimator is consistent for $\mathrm{AOI}_l$ as $K$ increases, while requiring only vector-Jacobian products rather than the full Jacobian. In our diagnostic figures, the plotted AOI is computed using this randomized estimator.

After estimating AOI, we compute answer-observable efficiency (AOE) by normalizing the estimated answer-observable influence by the visual update magnitude:
\[
    \widehat{\mathrm{AOE}}_l
    =
    \frac{
    \widehat{\mathrm{AOI}}_l
    }{
    \|\Delta V_l\|_F + \epsilon
    } .
\]
For visualization across layers, we further normalize AOE within each model and dataset:
\[
    \widetilde{\mathrm{AOE}}_l
    =
    \frac{
    \widehat{\mathrm{AOE}}_l
    }{
    \max_j \widehat{\mathrm{AOE}}_j + \epsilon
    } .
\]
\begin{table*}[!t]
    \centering
    \caption{
    Llava-Next comparison across multimodal benchmarks. 
    We report benchmark scores and average performance retention relative to the vanilla model. 
    }
    \label{tab:llava_results}
    \resizebox{\textwidth}{!}{
    \begin{tabular}{lllccccccccccc}
        \toprule
        Model & Method & TFLOPs & GQA & TextVQA & MME & MMB & MMMU & POPE & SQA & AI2D & OCRB & VizWiz & Avg. Ret. \\
        \midrule
        \rowcolor{gray!15} LLaVA-NeXT & Vanilla & 100\% & 64.30 & 61.36 & 1851.44 & 67.10 & 36.00 & 87.61 & 70.15 & 65.35 & 52.20 & 60.75 & 100.0 \\
        
         & VTW (K=16) & 51\% & 55.41 & 47.71 & 1852.56 & 66.75 & 35.56 & 87.41 & \textcolor{red}{70.00} & \textcolor{red}{65.38} & 6.90 & 54.86 & 86.5 \\

         & ShortV (N=20) & 51\% & 62.83 & 56.50 & \textcolor{red}{1854.51} & 66.92 & 35.89 & 87.38 & 69.26 & 64.48 & 41.40 & 57.09 & 96.0 \\

         & VSkip\textsuperscript{+} (N=20)& 74\% & \textcolor{red}{63.31} & 59.74 & 1759.67 & 67.01 & 36.22 & 87.35 & 69.83 & \textcolor{blue!60!black}{65.25} & 48.80 & \textcolor{blue!60!black}{59.35} & 98.2 \\
         
         & V2Drop & 70\% & 62.58 & 59.83 & \textcolor{blue!60!black}{1853.23} & \textcolor{red}{67.25} & 36.00 & \textcolor{blue!60!black}{87.98} & 69.92 & 64.75 & \textcolor{blue!60!black}{50.30} & 56.83 & 98.4 \\

         & APET & 70\% & 62.97 & \textcolor{blue!60!black}{59.98} & 1840.33 & \textcolor{blue!60!black}{67.18} & \textcolor{red}{37.11} & 87.79 & 69.71 & 65.01 & \textcolor{red}{50.60} & \textcolor{red}{60.72} & \textcolor{red}{99.4} \\

         & Ours (N=20) & 67\% & \textcolor{blue!60!black}{63.01} & \textcolor{red}{60.65} & 1797.57 & 66.75 & \textcolor{blue!60!black}{36.78} & \textcolor{red}{88.31} & \textcolor{blue!60!black}{69.96} & 65.03 & \textcolor{red}{50.60} & 57.99 & \textcolor{blue!60!black}{98.8} \\

        \bottomrule
    \end{tabular}}
\end{table*}

\afterpage{%
\begin{table*}[!t]
    \centering
    \caption{
    Sensitivity to the operator-skipping budget across the four evaluated MLLMs. Avg. Ret. is the mean retention relative to the vanilla model.
    }
    \label{tab:app_budget}
    \resizebox{\textwidth}{!}{%
    \begin{tabular}{lccccccccccc}
        \toprule
        Budget & GQA & TextVQA & MME & MMB & MMMU & POPE & SQA & AI2D & OCRB & VizWiz & Avg. Ret. \\
        \midrule
        \rowcolor{gray!15}\multicolumn{12}{c}{\textit{LLaVA-1.5-7B (32 Layers)}} \\
        0 & \textbf{61.94} & \textbf{58.21} & 1866.15 & 64.18 & \textbf{36.11} & 85.94 & 69.46 & 55.18 & \textbf{31.50} & \textbf{54.09} & 100.0 \\
        8 & 61.14 & 57.18 & \textbf{1878.10} & 64.18 & 35.67 & 86.43 & \textbf{69.61} & \textbf{55.51} & 31.40 & 53.23 & 99.6 \\
        12 & 61.10 & 57.12 & 1874.85 & 64.26 & 35.44 & 86.82 & 69.36 & 55.34 & 31.00 & 52.73 & 99.3 \\
        16 & 61.02 & 56.30 & 1856.99 & 64.18 & 35.89 & 86.93 & 69.36 & 55.18 & 31.30 & 52.45 & 99.1 \\
        20 & 60.84 & 56.40 & 1841.68 & 64.52 & 35.67 & \textbf{87.26} & 69.16 & 55.18 & 30.20 & 52.75 & 98.8 \\
        24 & 60.71 & 56.22 & 1797.31 & 64.61 & 35.78 & 87.22 & 69.16 & 55.28 & 29.60 & 53.46 & 98.5 \\
        28 & 60.23 & 55.61 & 1784.78 & \textbf{65.38} & 34.67 & 86.94 & 69.16 & 54.53 & 28.30 & 53.06 & 97.4 \\
        32 & 57.89 & 53.80 & 1728.08 & 62.03 & 35.67 & 85.01 & 69.21 & 52.46 & 25.10 & 52.76 & 94.5 \\
        \midrule
        \rowcolor{gray!15}\multicolumn{12}{c}{\textit{LLaVA-NeXT-7B (32 Layers)}} \\
        0 & \textbf{64.30} & 61.36 & \textbf{1851.44} & \textbf{67.10} & 36.00 & 87.61 & \textbf{70.15} & 65.35 & \textbf{52.20} & 60.75 & 100.0 \\
        8 & 63.75 & \textbf{61.37} & 1845.83 & \textbf{67.10} & 36.33 & 87.54 & \textbf{70.15} & \textbf{65.48} & 51.40 & \textbf{61.05} & 99.9 \\
        12 & 63.55 & 61.08 & 1837.72 & 67.01 & 35.67 & 87.57 & 70.10 & 65.32 & 51.30 & 60.06 & 99.4 \\
        16 & 63.36 & 60.77 & 1832.66 & \textbf{67.10} & 36.33 & 87.57 & 69.96 & 65.19 & 51.20 & 59.45 & 99.3 \\
        20 & 63.01 & 60.65 & 1797.57 & 66.75 & \textbf{36.78} & \textbf{88.31} & 69.96 & 65.03 & 50.60 & 57.99 & 98.8 \\
        24 & 62.30 & 58.36 & 1777.99 & 66.15 & 36.00 & 88.20 & 69.26 & 63.76 & 45.70 & 58.21 & 96.7 \\
        28 & 61.75 & 56.49 & 1744.20 & 64.18 & 35.00 & 86.87 & 68.67 & 63.02 & 39.50 & 59.73 & 94.3 \\
        32 & 59.64 & 50.67 & 1616.91 & 59.54 & 34.00 & 85.26 & 66.48 & 61.27 & 26.30 & 58.67 & 87.9 \\
        \midrule
        \rowcolor{gray!15}\multicolumn{12}{c}{\textit{Qwen2.5-VL-7B (28 Layers)}} \\
        0 & \textbf{60.40} & \textbf{77.77} & \textbf{2515.82} & \textbf{83.25} & \textbf{50.00} & \textbf{87.62} & \textbf{87.51} & 82.42 & \textbf{84.10} & \textbf{70.81} & 100.0 \\
        8 & 59.85 & 76.62 & 2346.15 & 82.56 & 49.11 & 87.29 & 87.41 & \textbf{82.58} & 81.80 & 68.25 & 98.2 \\
        12 & 59.60 & 74.52 & 2343.01 & 82.65 & 49.33 & 87.08 & 87.06 & 82.22 & 80.50 & 66.77 & 97.4 \\
        16 & 59.38 & 73.41 & 2281.94 & 81.70 & 48.22 & 86.43 & 86.91 & 81.41 & 78.40 & 66.16 & 96.1 \\
        20 & 59.29 & 72.02 & 2263.38 & 80.67 & 46.89 & 87.32 & 85.52 & 79.95 & 76.90 & 65.48 & 95.0 \\
        24 & 58.75 & 70.28 & 2159.23 & 79.38 & 46.56 & 86.52 & 82.94 & 78.82 & 70.60 & 64.84 & 92.7 \\
        28 & 57.27 & 54.75 & 1884.70 & 69.07 & 39.33 & 86.29 & 77.94 & 62.08 & 42.30 & 54.81 & 79.2 \\
        \midrule
        \rowcolor{gray!15}\multicolumn{12}{c}{\textit{Qwen3-VL-8B (36 Layers)}} \\
        0 & \textbf{61.60} & \textbf{80.07} & \textbf{2390.38} & \textbf{84.79} & 51.33 & 89.13 & 94.40 & \textbf{83.78} & \textbf{82.80} & 69.37 & 100.0 \\
        12 & 60.98 & 79.93 & 2352.81 & 84.62 & 52.44 & 88.83 & \textbf{94.70} & 83.65 & 81.90 & 70.78 & 100.0 \\
        16 & 61.04 & 79.64 & 2356.06 & 84.45 & \textbf{53.67} & 88.76 & \textbf{94.70} & 83.65 & 81.40 & 70.12 & 100.0 \\
        20 & 60.70 & 78.45 & 2336.01 & 83.08 & 52.89 & \textbf{89.43} & 94.35 & 82.55 & 80.40 & \textbf{71.70} & 99.5 \\
        24 & 60.24 & 77.01 & 2273.65 & 82.05 & 53.00 & 89.31 & 90.53 & 80.18 & 77.90 & 69.77 & 97.6 \\
        28 & 59.97 & 75.58 & 2158.87 & 80.93 & 51.00 & 88.42 & 85.72 & 76.39 & 71.20 & 66.86 & 94.0 \\
        32 & 59.59 & 72.67 & 1909.52 & 78.01 & 49.44 & 88.42 & 85.72 & 76.39 & 71.20 & 66.86 & 91.9 \\
        36 & 58.51 & 69.91 & 1707.86 & 75.34 & 48.22 & 87.42 & 81.85 & 72.38 & 63.80 & 63.77 & 87.7 \\
        \bottomrule
    \end{tabular}%
    }
\end{table*}
}

Therefore, VUM measures how much visual-token states move, AOI estimates how much this movement reaches the final prompt-token hidden state, and AOE measures the answer-observable influence induced per unit visual update.

In implementation, we first run the original model forward pass and cache $H_v^l$, $H_v^{l+1}$, and $h_{\mathrm{last}}$ for each layer. For each layer $l$, we sample $K$ random probes $\{r_k\}_{k=1}^K$, compute the scalar $r_k^\top h_{\mathrm{last}}$, backpropagate it to obtain the gradient with respect to $H_v^{l+1}$, and take its Frobenius inner product with $\Delta V_l$. Model parameters are kept frozen, and no ground-truth answer labels are used. This makes the diagnostic label-free and applicable to arbitrary multimodal prompts.

\section{Experiments on LLaVA-NeXT}
\label{app:llava_next}

To further test architectural generality, we additionally evaluate on LLaVA-NeXT. Compared with LLaVA-1.5, LLaVA-NeXT uses stronger visual processing and is commonly evaluated under higher-resolution or multi-image settings, making visual-token computation more expensive. This provides a useful stress test for operator-aware skipping because high-resolution visual inputs amplify the cost of repeatedly updating visual-token representations inside the LLM backbone.

Table~\ref{tab:llava_results} shows that our method remains effective in this higher-resolution setting. With $67\%$ TFLOPs, our method preserves $98.8\%$ average performance and achieves the best results among non-vanilla methods on TextVQA and POPE, while also ranking second on GQA, MMMU, and SQA. It also matches the best OCRBench score. These results indicate that the operator-level redundancy identified in the main experiments is not limited to the three primary backbones, but also appears in LLaVA-NeXT. Even when the visual input is more expensive to process, selectively skipping redundant visual operators can reduce computation while maintaining strong benchmark performance.

\section{Calibration Sample Size Study}
\label{app:calibration_size}

We study how the number of calibration samples from each benchmark affects the analysis decisions made by our method. For each of the seven calibration benchmarks used in the main experiments, we randomly sample $8$, $16$, or $20$ examples and estimate the layer-wise operator risks for LLaVA-1.5-7B, Qwen2.5-VL-7B, and Qwen3-VL-8B. The resulting risk profile determines the selected layers and their operator-skipping policies, and the same policy is then evaluated on the full benchmark suite.

Table~\ref{tab:app_calibration_size} reports the average retention under different layer budgets. The goal is to test whether using more calibration samples leads to a substantially more reliable policy. Since calibration is used only for risk estimation and does not require labels or model training, a small sample size is desirable if it already supports stable policy selection.

\begin{center}
\begin{minipage}{\columnwidth}
\captionsetup{hypcap=false}
\captionof{table}{Effect of calibration sample size under different layer budgets. Each setting samples the same number of examples from each of seven calibration benchmarks.}
\label{tab:app_calibration_size}
\resizebox{\columnwidth}{!}{
\begin{tabular}{cccccccccc}
\toprule
\begin{tabular}[c]{@{}c@{}}Model\end{tabular} & \begin{tabular}[c]{@{}c@{}}Samples\\Per Benchmark\end{tabular} & \multicolumn{8}{c}{Avg. Ret. under Layer Budget} \\
\cmidrule(lr){3-10}
 & & 8 & 12 & 16 & 20 & 24 & 28 & 32 & 36 \\
\midrule
 & 8  & 99.58 & 99.25 & 99.15 & 98.76 & 98.46 & 97.37 & 94.48 & -- \\
LLaVA-1.5-7B & 16 & 99.62 & 99.37 & 99.28 & 98.81 & 98.52 & 97.49 & 94.32 & -- \\
 & 20 & 99.74 & 99.55 & 99.52 & 99.01 & 98.57 & 97.60 & 94.08 & -- \\
\midrule
 & 8  & 98.16 & 97.42 & 96.14 & 94.97 & 92.66 & 79.23 & -- & -- \\
Qwen2.5-VL-7B & 16 & 98.18 & 97.51 & 96.17 & 95.10 & 92.68 & 79.26 & -- & -- \\
 & 20 & 97.92 & 97.76 & 96.65 & 95.19 & 92.63 & 79.31 & -- & -- \\
\midrule
 & 8  & -- & 99.98 & 100.04 & 99.45 & 97.55 & 95.73 & 91.93 & 87.67 \\
Qwen3-VL-8B & 16 & -- & 99.99 & 100.01 & 99.32 & 97.56 & 94.91 & 91.49 & 86.75 \\
 & 20 & -- & 100.06 & 100.10 & 99.15 & 97.16 & 93.60 & 88.42 & 84.22 \\
\bottomrule
\end{tabular}}
\end{minipage}
\end{center}

The results show that increasing the number of calibration samples can bring improvements at some layer budgets, but the gains are not consistently large or monotonic. For example, more samples slightly improve several LLaVA-1.5-7B settings, while the trends for Qwen2.5-VL-7B and Qwen3-VL-8B vary across budgets. This suggests that $8$ samples per benchmark are already sufficient to support the risk analysis used by our method. Adding more samples can provide stronger evidence in some cases, but it is not always necessary for obtaining stable operator-skipping decisions.

\begin{figure*}[t]
    \centering
    \includegraphics[width=\textwidth]{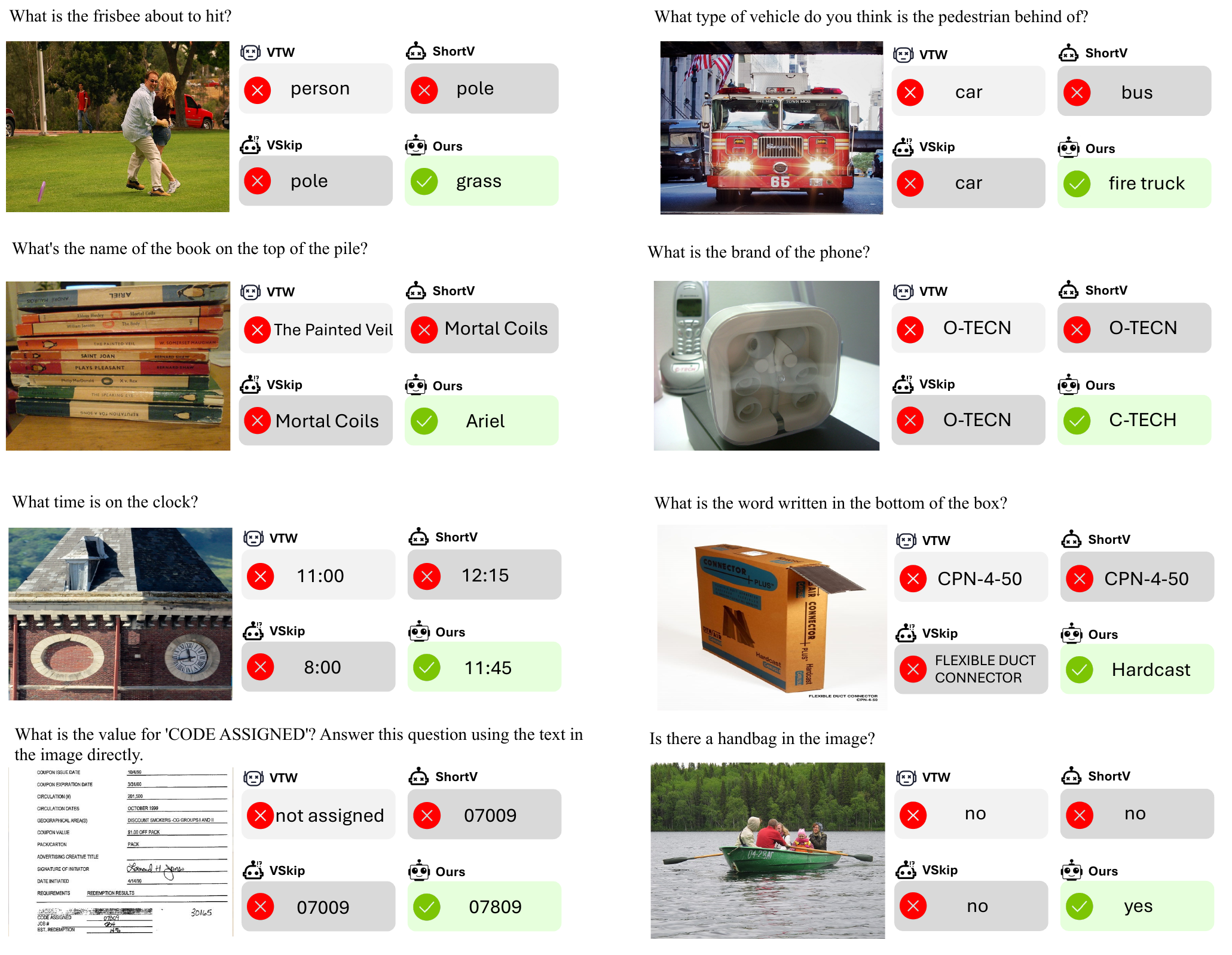}
    \caption{Additional qualitative examples of operator-aware visual-token skipping.}
    \label{fig:app_more_examples}
\end{figure*}

\section{Ablation on Risk-Based Layer Allocation}
\label{app:last20_ablation}

We compare our budgeted layer allocation with a simple last-layer heuristic. As described in Section~\ref{sec:operator-aware-skipping}, our policy first ranks layers by the summed single-operator risk $\mathcal{R}_{\mathcal A}(l)+\mathcal{R}_{\mathcal F}(l)$ and allocates the skipping budget to the lowest-risk layers. This design assumes that layers with smaller summed risk can safely discard their non-dominant visual operator earlier, whereas layers with larger summed risk should be postponed because even their non-dominant operator may still affect the output.

\begin{center}
\begin{minipage}{\columnwidth}
\captionsetup{hypcap=false}
\captionof{table}{Ablation against a heuristic that skips visual computation in the last 20 layers on Qwen3-VL-8B.}
\label{tab:app_last20}
\resizebox{\columnwidth}{!}{
\begin{tabular}{lccccccc}
\toprule
Method & GQA & TextVQA & MME & MMB & SQA & AI2D & OCRB \\
\midrule
Last-20 heuristic & 60.57 & 77.40 & 2282.61 & 82.47 & 93.01 & 81.44 & 78.80 \\
Ours (N=20) & 60.70 & 78.45 & 2336.01 & 83.08 & 94.35 & 82.55 & 80.40 \\
\bottomrule
\end{tabular}}
\end{minipage}
\end{center}

This ablation isolates the effect of this layer-allocation rule. For both methods, we keep the per-layer operation type fixed; each selected layer uses the same skip-operator decision or the same frozen update determined by the operator-risk profile. The only difference is which layers are selected. Our method selects the 20 layers with the smallest $\mathcal{R}_{\mathcal A}(l)+\mathcal{R}_{\mathcal F}(l)$, while the Last-20 heuristic directly selects the final 20 Transformer layers regardless of their risk scores. Table~\ref{tab:app_last20} reports the Qwen3-VL-8B comparison, testing whether risk-based allocation is preferable to choosing late layers purely by position.

The results support our hypothesis. With the same number of selected layers and the same per-layer operator actions, risk-based allocation consistently outperforms the Last-20 heuristic across all reported benchmarks. This indicates that small-$\mathcal{R}_{\mathcal A}+\mathcal{R}_{\mathcal F}$ layers are indeed safer places to discard non-dominant visual operators, while simply choosing the final layers may remove useful visual computation whose operator risk remains high.

\section{Different Budget of Our Method}
\label{app:budget}

Table~\ref{tab:app_budget} studies how the operator-skipping budget affects performance across four MLLM backbones. Overall, the results show that our operator-aware policy is not tied to a particular architecture, LLaVA-1.5, LLaVA-NeXT, Qwen2.5-VL, and Qwen3-VL all maintain high average retention under moderate budgets, despite their different visual encoders, language backbones, and dependency stacks. In particular, small and medium budgets usually preserve nearly all vanilla performance, and several individual benchmarks even improve over the budget-0 setting. This suggests that the skipped operators are often answer-silent rather than uniformly useful, supporting the reliability of the proposed answer-observable selection criterion.

The budget trend also reveals a consistent efficiency--accuracy pattern. As the budget increases, average retention decreases smoothly rather than collapsing abruptly, indicating that the policy removes redundant visual computation in a controlled manner. This behavior is especially important for practical deployment, users can choose conservative budgets when accuracy is the priority, or larger budgets when higher efficiency is needed. Across all models, moderate budgets retain strong performance, while aggressive settings expose the trade-off when too much visual computation is skipped. These results show that operator-level visual skipping is broadly applicable and robust under different layer-level budgets.



\end{document}